\documentclass[a4paper,11pt,numbers]{elsarticle}

\usepackage{graphicx}
\usepackage[fleqn]{amsmath}
\usepackage{amsfonts}
\usepackage{amssymb}
\usepackage{bm}

\usepackage{float}
\usepackage[dvips]{epsfig}
\usepackage{epstopdf}
\usepackage{rotating}
\usepackage{tikz}
\usetikzlibrary{shapes,positioning,intersections,quotes,bayesnet}
\usetikzlibrary{decorations.pathmorphing}
\usetikzlibrary{decorations.markings}
\usetikzlibrary{arrows.meta,bending}

\usepackage{setspace} 
\usepackage{lineno}   
\modulolinenumbers[5]

\usepackage{microtype}
\usepackage{url}
\usepackage{subfiles}
\usepackage{caption}
\usepackage{subcaption}
\usepackage{parskip}
\usepackage[section]{placeins}
\usepackage{xcolor}
\usepackage{framed}
\colorlet{shadecolor}{blue!20}
\usepackage{tabulary}
\usepackage[hidelinks]{hyperref}
\usepackage[capitalize,nameinlink]{cleveref}
\usepackage{flexisym}

\usepackage{multirow, bigdelim}

\newcommand{\ue}{\underline{e}}

\newcommand{\ux}{\underline{x}}

\newcommand{\uf}{\underline{f}}

\newcommand{\uu}{\underline{u}}
\newcommand{\uv}{\underline{v}}

\newcommand{\R}{\mathbb{R}}

\newcommand{\uth}{\underline{\theta}}

\graphicspath{{./Figures/}}

\journal{Mechanical Systems and Signal Processing}

\begin{document}
\journal{Mechanical Systems and Signal Processing}

\begin{frontmatter}

\title{Foundations of Population-Based SHM, Part IV: The Geometry of Spaces of Structures and their Feature Spaces}

\author[1]{G.\ Tsialiamanis\footnote{Corresponding Author: George Tsialiamanis (g.tsialiamanis@sheffield.ac.uk)}}
\author[2]{C.\ Mylonas}
\author[2]{E.\ Chatzi}
\author[1]{N.\ Dervilis}
\author[1]{D.J.\ Wagg}
\author[1]{K.\ Worden}
\address[1]{Dynamics Research Group, Department of Mechanical Engineering, University of Sheffield \\ Mappin Street, Sheffield S1 3JD}
\address[2]{ETH Zurich, Institute of Structural Engineering, Stefano-Franscini-Platz 5, 8093 Zurich}

\begin{abstract}
One of the requirements of the population-based approach to Structural Health Monitoring (SHM) proposed in the earlier papers in this sequence, is
that structures be represented by points in an abstract space. Furthermore, these spaces should be metric spaces in a loose sense; i.e.\ there
should be some measure of distance applicable to pairs of points; similar structures should then be ‘close’ in the metric. However, this geometrical construction
is not enough for the framing of problems in data-based SHM, as it leaves undefined the notion of feature spaces. Interpreting the feature values on a
structure-by-structure basis as a type of field over the space of structures, it seems sensible to borrow an idea from modern theoretical physics, and
define feature assignments as sections in a vector bundle over the structure space. With this idea in place, one can interpret the effect of environmental
and operational variations as gauge degrees of freedom, as in modern gauge field theories. One can then regard data normalisation procedures like
cointegration as gauge-fixing operations. This paper will discuss the various geometrical structures required for an abstract theory of feature spaces in
SHM, and will draw analogies with how these structures have shown their power in modern physics.

Having motivated a number of problems in Population-Based SHM (PBSHM) in geometrical terms, it remains to show how these problems might be solved. In the
second part of the paper, the problem of determining the {\em normal condition cross section} of a feature bundle is addressed. The solution is provided by
the application of {\em Graph Neural Networks} (GNN), a versatile non-Euclidean machine learning algorithm which is not restricted to inputs and outputs
from vector spaces. In particular, the algorithm is well suited to operating directly on the sort of graph structures which are an important part of the
proposed framework for PBSHM. The solution of the normal section problem is demonstrated for a heterogeneous population of truss structures for which the
feature of interest is the first natural frequency. The GNN approach is shown to not only solve the normal section problem, but also to accommodate
varying temperatures across the population; it thus provides a means of data normalisation.
\end{abstract}

	
\end{frontmatter}

\graphicspath{{./Figures/}}

\section{Introduction}
\label{sec:intro}

This paper is the fourth in a sequence devoted to introducing foundations for a new discipline of {\em Population-based Structural Health Monitoring} (PBSHM)
\cite{PBSHMMSSP1,PBSHMMSSP2,PBSHMMSSP3}. The aim of the new technology is to facilitate the principled transfer of information between disparate structures,
specifically for SHM diagnostic purposes. In the first paper, the idea of a {\em homogenenous population} was introduced, and the concept of the {\em form}
appeared as a means of representing populations of nominally-identical structures \cite{PBSHMMSSP1}. The next papers in the sequence introduced the more
challenging {\em heterogeneous population}, which is formed of disparate structures \cite{PBSHMMSSP2,PBSHMMSSP3}. In order to impose
some mathematical order on the heterogeneous populations, the Irreducible Element (IE) model and its associated Atributed Graph (AG) were introduced as abstract
representations of structures, with the population then taking the form of a complex network \cite{PBSHMMSSP2}. In this framework, the physical structures of
interest appear as points in a (mathematically) structured set, which has metric properties that allow a judgement of the closeness of resemblance of the physical
structures. This metric is a key element in deciding whether two structures are `close' enough to allow the transfer of diagnostic information or capability
\cite{PBSHMMSSP3}.

In reality, the `transfer' of information will take the form of maps and associations between the feature spaces associated with the SHM problems specified
for the structures. The feature spaces themselves will usually be vector spaces. In abstract terms, one might think of the population itself as having a
total feature space, which is the union of all those of the individuals. In rough terms, one can think of this object as a type of vector bundle \cite{Schutz,Hamilton}
over the space of structures, with `transfer' being a map within the bundle space. The first aim of the current paper is to look at whether this level of
abstraction is sensible -- or even possible -- and to speculate on whether it might bring practical benefits for PBSHM.

Vector bundles, or more generally fibre bundles, most often appear in algebraic or differential geometry, and are often constructed from {\em differentiable
manifolds}, which are spaces where notions of smoothness and differentiability -- the ability to meaningfully do calculus -- are important. The `spaces' of
structures mentioned earlier, the complex networks of attributed graphs, do not have such properties of smoothness; however, in order to explore the possibility
of using fibre bundles in PBSHM, the discussion here will begin with some problems in structural dynamics in which the spaces of structures are indeed
manifolds. This will lead to the idea of feature spaces as bundles above manifolds, and then finally to the general mathematical structures of interest in PBSHM.

It is argued that the bundle representation also accommodates other aspects of traditional SHM, like {\em confounding influences}. These influences arise when SHM
is implemented via {\em change} or {\em novelty detection} i.e.\ signalling when feature data are inconsistent with some previously-learned model of normal
condition. For example, an SHM system for a bridge might produce a false alarm if the bridge data change because of some benign cause like a change in temperature
or traffic loading \cite{Alampalli,FarrarI40,Peeters}. The solution to the problem is {\em data normalisation}, whereby the effects of benign changes are projected
out from the SHM feature data \cite{Sohn}. It is argued in the current paper that confounding influences can be considered as analogous to the gauge variations seen
in modern field theory, and that data normalisation can be regarded as a form of gauge fixing \cite{Hamilton}.

Having given a geometrical context for PBSHM problems, it remains to suggest how they might be solved, and this is the subject of the second part of the paper. In
particular a specific geometrical problem -- the {\em normal condition cross section} or {\em normal section} problem -- is addressed. This is the problem of
determining the normal condition features for an entire population when measured data are only available from a subset of the population. The solution proposed
here makes use of a modern non-Euclidean machine learning algorithm -- the {\em Graph Neural Network} (GNN) -- which can operate on graph structures directly, rather
than first embedding them in some real vector space. This algorithm is ideally suited to PBSHM, where the structures of interest are represented by a complex
network of attributed graphs \cite{PBSHMMSSP2}

The GNN approach is demonstrated on the normal section problem for a heterogeneous population of truss structures, where the SHM feature of interest is the first natural
frequency of the trusses. It is shown to interpolate effectively across the population, even when temperature variations are present in the data of individual trusses.
Because of its ability to accommodate temperature variations, the approach also solves the data normalisation problem associated with the population.

The layout of the paper is as follows. Section Two will provide basic introductions to fibre bundles, while Section Three will provide some examples of how
fibre bundles might arise in structural dynamics. Section Four introduces the idea of feature spaces as vector bundles and Section Five extends the discussion to
include confounding influences and data normalisation. Section Six discusses how the feature bundles might arise in the context of spaces of structures for PBSHM.
Having motivated the geometrical viewpoint of problems, the paper moves on to possible solutions. Section Seven discusses how traditional machine learning has
difficulties in analysing input and output data which do not live in vector spaces, and outlines some of the specific difficulties encountered when the objects of
interest are graphs. Section Eight introduces the idea of Graph Networks (GNs), which can operate directly on graph objects, and gives the background theory on
how such networks can be trained; the representation of certain functions in the framwework by neural networks then motivates the Graph Neural Network (GNN)
algorithm. In Section Nine, the GNN algorithm is demonstrated on the normal section problem for a heterogeneous population of truss structures. The paper ends with
sections of discussion and short conclusions.

Throughout this paper, underlines will denote vectors, while square brackets will denote matrices.

\section{Fibre bundles}
\label{sec:bundles}

The basic idea of the fibre bundle is explained well in \cite{Schutz}; however, only the basic definitions are covered. Much deeper mathematical coverage can be
found in a number of `classic' texts \cite{Steenrod,Husemoller,Kobayashi}. As many of the ideas discussed in this paper are motivated by the mathematical physics of
{\em gauge field theories}, the interested reader can find fundamentals explained in the survey paper \cite{Eguchi}, and text \cite{Hamilton}. The theory of fibre
bundles is founded on the idea of a {\em differential manifold}; further interested readers may refer to \cite{Schutz}. Less
background is provided on manifolds because they do appear more regularly in the SHM literature; this is because they appear at the interface of dynamics and
machine learning. Interesting discussions in terms of (nonlinear) dynamics can be found in \cite{Abraham}, based on foundational work in \cite{Reeb,Synge,Mackey}.
In the context of machine learning, the class of algorithms termed {\em manifold learning} \cite{Izenman} have proved powerful, particularly in terms of data
visualisation \cite{McInnes}.

\subsection{Fibre bundles - basic definitions}
\label{subsec:def_bundles}

A fibre bundle is fundamentally composed of two objects: a {\em base manifold} $M$ (e.g.\ spacetime), and a {\em total manifold} $E$, which essentially
contains the `fields on $M$'. In order that movement on the base manifold induces movement in the total manifold, the two manifolds need to be glued
together, and this is accomplished by means of a projection $\pi$ which is a map from $E$ to $M$. Now, thinking in terms of the number of components one
needs to specify for a vector field, it is clear that $E$ must have a {\em higher dimension} than $M$. This in turn means that the inverse map $\pi^{-1}(x)$
at any point $x \in M$ must be multi-valued. The first requirement in defining a fibre bundle $\pi: E \longrightarrow M$ is that all the sets
$\pi^{-1}(x) = F_x \in E$ are all homeomeorphic to each other; these sets are denoted $F$ and termed {\em the fibre above $x$}.

\begin{figure}[htbp!]
    \centering
     \begin{tikzpicture}
     \definecolor{blue1}{RGB}{93, 143, 218}
     \definecolor{teal}{RGB}{100, 225, 225}
     \draw[line width=0.2mm, blue1] (0.0, 0.0) to[out=30, in=150] (3.0, 0.5) to[out=330, in=210] (6.0, 1.0);
     \draw[line width=0.2mm, blue1] (6.0, 1.0) to[] (7.0, 3.0);
     \draw[line width=0.2mm, blue1] (1.0, 2.0) to[out=30, in=150] (3.0, 2.5) to[out=330, in=210] (7.0, 3.0);
     \draw[line width=0.2mm, blue1] (1.0, 2.0) to[] (0.0, 0.0);
     
     
     \draw[line width=0.2mm, blue1] (0.0, 4.0) to[] (0.0, 7.0);
     \draw[line width=0.2mm, blue1] (6.0, 4.5) to[] (6.0, 7.5);
     
     \draw[line width=0.2mm, blue1] (0.0, 7.0) to[] (1.0, 8.5);
     \draw[line width=0.2mm, blue1] (6.0, 7.5) to[] (7.0, 9.0);
     
     \draw[line width=0.2mm, blue1] (1.0, 8.5) to[out=60, in=150] (3.0, 8.75) to[out=330, in=240] (5.0, 8.75) to[out=60, in=130] (7.0, 9.0);
     
     \draw[line width=0.2mm, blue1] (6.0, 4.5) to[] (7.0, 6.0);
     \draw[line width=0.2mm, blue1] (7.0, 6.0) to[] (7.0, 9.0);
     
     \node[] (M) at (7.0, 2.0) {$M$};
     
     \node[] (E) at (7.6, 7.0) {$E$};
     
     \node[circle,color=blue1, fill=blue1, inner sep=0pt,minimum size=3pt] (fb1) at (2.0, 4.75) {};
     \node[circle,color=blue1, fill=blue1, inner sep=0pt,minimum size=3pt] (fb2) at (2.0, 7.75) {};
     \draw[line width=0.5mm, blue1] (fb1) to[] (fb2);
     \node[] (F) at (2.2, 5.75) {$F$};

     \node[circle,color=black, fill=blue1, inner sep=0pt,minimum size=3pt] (x) at (2.0, 1.0) {};
     \draw[-{>[scale=2.5, length=2, width=3]}, line width=0.4mm, color=teal] (x) to (fb1);
     
     \node[] (x_) at (2.3, 1.0) {$x$};
     \node[] (phi) at (2.5, 2.25) {$\pi^{-1}$};
     
     \draw[line width=0.2mm, blue1] (0.0, 4.0) to[out=60, in=150] (2.0, 4.25) to[out=330, in=240] (4.0, 4.25) to[out=60, in=130] (6.0, 4.5);
     \draw[line width=0.2mm, blue1] (0.0, 7.0) to[out=60, in=150] (2.0, 7.25) to[out=330, in=240] (4.0, 7.25) to[out=60, in=130] (6.0, 7.5);
     
     \end{tikzpicture}
    \caption{Schematic of basic objects in a fibre bundle.}
    \label{fig:l9_def_bundle}
\end{figure}
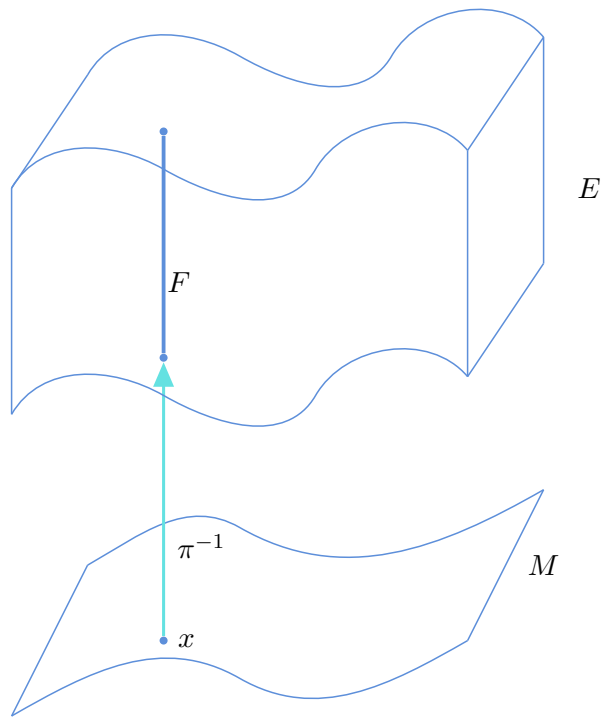

So far, the fibre bundle could be defined as the set of objects $\{ M, E, \pi, F \}$ as depicted in Figure \ref{fig:l9_def_bundle}; however, one needs a little
more precision. If there is a copy of $F$ above each point of $M$, a simple way to define $E$ would be as the Cartesian product $M \times F$, then the
projection is simply $\pi(x,f) = x$ and one obtains $\pi^{-1}(x) = F$, exactly as required. However, there is no new structure here, it really is just a
Cartesian product. One can allow $E$ to have much more general structures by using local coordinates in the same way that allows the definition of a manifold in the
first place \cite{Schutz}. Suppose that one is working within a specific coordinate domain $U$ on $M$, there is then the local homeomorphism
into $\R^n$ (where $n$ is the dimension of $M$). To mimic all of the appropriate flat-space vector calculus, one can assume that $\pi^{-1}(U) = U \times F$ locally,
but as in the case of $M$ itself, one does not have to assume that the single homeomorphism extends across all $M$ e.g.\ $\pi^{-1}(M) \neq M \times F$.

\begin{figure}[htbp!]
    \centering
     \begin{tikzpicture}[scale=0.9, every node/.style={scale=0.9}]
     \definecolor{blue1}{RGB}{93, 143, 218}
     \definecolor{blue2}{RGB}{66, 170, 178}
     \definecolor{teal}{RGB}{100, 225, 225}
     
     \draw[line width=0.2mm, blue1] (0.0, 0.0) to[out=30, in=150] (3.0, 0.5) to[out=330, in=210] (6.0, 1.0);
     
     \draw[line width=0.2mm, blue1] (0.0, 4.0) to[out=60, in=150] (2.0, 4.25) to[out=330, in=240] (4.0, 4.25) to[out=60, in=130] (6.0, 4.5);
     
     \draw[line width=0.2mm, blue1] (0.0, 4.0) to[] (0.0, 7.0);
     \draw[line width=0.2mm, blue1] (6.0, 4.5) to[] (6.0, 7.5);

     \node[] (M) at (6.5, 1.0) {$M$};
     
     \node[] (E) at (6.4, 7.0) {$E$};

     
     \node[] (x_) at (2.0, 2.0) {$\pi^{-1}$};
     
     \draw[-{>[scale=2.5, length=2, width=3]}, line width=0.4mm, color=teal] (2.0, 2.3) to[] (2.0, 3.4);
     
     \draw[line width=0.2mm, blue1] (1.5, 0.68) to[] (1.5, 7.45);
     \draw[line width=0.2mm, blue1] (2.5, 0.7) to[] (2.5, 7.0);

     \draw[line width=0.2mm, blue1] (0.0, 7.0) to[out=60, in=150] (2.0, 7.25) to[out=330, in=240] (4.0, 7.25) to[out=60, in=130] (6.0, 7.5);
     \node[] (U_i) at (2.2, 0.5) {$U_{i}$};
     
     
     \draw[-{>[scale=2.5, length=2, width=3]}, line width=0.4mm, color=teal] (2.0, 6.0) to[out=-20, in=130] (5.0, 5.5) to[out=310, in=180] (8.0, 5.0);
     
     \node[] (phi_1) at (5.0, 6.0) {$\varphi_{i}$};
     
     
     \draw[line width=0.3mm, blue2] (8.3, 5.0) to[] (8.3, 3.0);
     \draw[line width=0.3mm, blue2] (11.3, 5.0) to[] (11.3, 3.0);
     \draw[line width=0.3mm, blue2] (8.3, 5.0) to[] (11.3, 5.0);
     \draw[line width=0.3mm, blue2] (8.3, 3.0) to[] (11.3, 3.0);
     
     \node[] (U_i2) at (9.8, 2.7) {$U_{i}$};
     \node[] (F) at (11.8, 4.0) {$F$};
     
     \fill[blue2, fill opacity=0.4] (1.5, 7.45) to[out=-8, in=160] (2.5, 7.0) to ++(0.0, -3.0) to[out=160, in=-10] (1.5, 4.45);
     
     \fill[blue2, fill opacity=0.4] (8.3, 5.0) to (8.3, 3.0) to (11.3, 3.0) to (11.3, 5.0);
     
     \draw[line width=0.5mm, blue2] (1.5, 0.68) to[out=12, in=168] (2.5, 0.7);
     
     \end{tikzpicture}
    \caption{Local triviality property in a fibre bundle.}
    \label{fig:l9_loc_triv}
\end{figure}
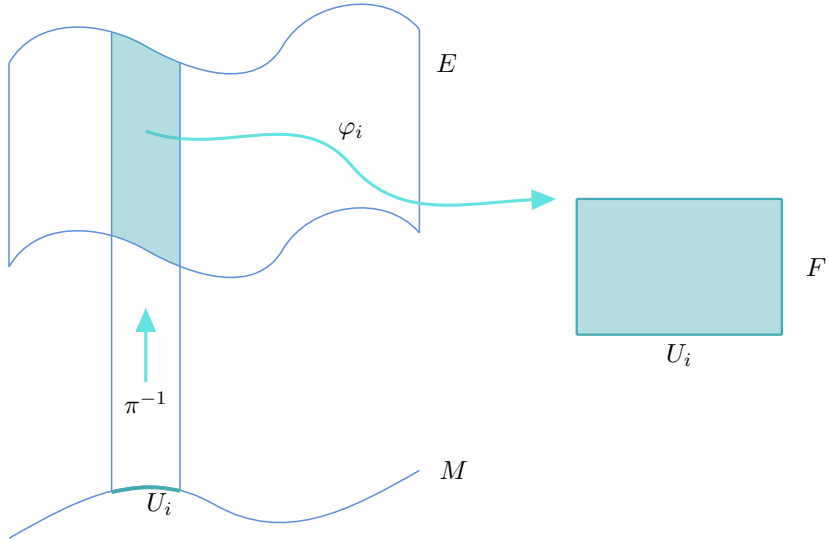

In much the same way that a manifold is only {\em locally homeomorphic} to $\R^n$ \cite{Schutz}, one can require that the total space
$E$ is only {\em locally homeomorphic} to the product space. One extends the definition of the bundle to the set of objects $\{ M, \{U_i\}, E, \{\varphi_i\}, \pi, F \}$
where $\{U_i\}$ is a coordinate atlas on $M$ and for each $U_i$ on $M$, $\varphi_i$ is a homeomorphism (Figure \ref{fig:l9_loc_triv}),

\begin{equation}
     \varphi_i : \pi^{-1}(U_i) \longrightarrow U_i \times F
\label{eq:l9_def_phi}
\end{equation}

This property is also referred to as {\em local triviality}.

In reality, things are a little more complicated. In much the same way that coordinate patches have to glue together in a particular way in order to respect
the topology of a manifold, via appropriately smooth changes of coordinates, the locally-trivial regions of the fibre bundle need to be glued together;
this is accomplished by means of {\em structure functions}. The structure functions do not play a role in the remainder of this paper, in order to find out
more, the reader can consult \cite{Steenrod,Eguchi}.

At this point, one can define the objects that correspond to `fields' in the theory. Supposing the base manifold to be spacetime, a vector field (for example),
would be the assignment of a single vector to each point in spacetime. If the fibre space is a vector space of the appropriate dimension, a vector field
is thus a map from points $x$ in the base manifold, to points in the fibres above $x$. Such maps $s(x) \in F$ are called {\em sections} and are defined by
the condition $\pi( s(x) ) = x$ or alternatively by $\pi \circ s = Id$. Clearly, this definition is too general as it stands, so it is always supplemented
by conditions on the map like smoothness of some degree. According to these ideas, a vector field is a section of the tangent bundle \cite{Schutz}.

When the fibre is a vector space, the origin of the space is a special point and thus generates a special section -- the {\em zero section} -- when the image
of $s(x)$ in each fibre is the origin \cite{Eguchi}.

\section{A Simple Illustration of a Fibre Bundle in Structural Dynamics}
\label{sec:ex_bundles}

Like manifolds in general, fibre bundles have been used extensively in the past in the mathematical theory of nonlinear dynamics \cite{Abraham} and in gauge theories
of particle physics \cite{Hamilton}. However, they have not really featured in engineering/structural dynamics at all. The purpose of this section will be to show that
some problems in structural dynamics can be framed in terms of bundles; in particular, some of the structures that would appear to be desirable from the point of view of
population-based Structural Health Monitoring (PBSHM) have abstract formulations in those terms \cite{PBSHMMSSP1,PBSHMMSSP2,PBSHMMSSP3}. One of the main aims of PBSHM is
to allow the transfer of SHM inferences between different structures, in situations where data for training machine learners are available for some structures -- {\em
source structures} -- but not others -- {\em target structures} \cite{PBSHMMSSP3}. A mathematical formulation of PBSHM can be defined in terms of an abstract
representation of structures, as discussed in \cite{PBSHMMSSP2,PBSHMMSSP3}. One way to build such a representation theory would be to have the structures of interest as {\em
points in a space of structures}; the feature data that are then used for inference can be envisaged as points in a bundle space over the space of structures.
As will be seen later, this may require quite a lot of generalisation, as the `space of structures' might not be a manifold.

To provide an illustration of the use of bundles, one can consider the simplest dynamical situation possible -- the Single-Degree-of-Freedom (SDOF) linear oscillator,
governed by the equation of motion,

\begin{equation}
     m \ddot{y} + c \dot{y} + k y = 0
\label{eq:l9_sdof}
\end{equation}
for the situation of unforced or free motion. This `structure' is uniquely fixed by the values of $m$, $c$ and $k$, so the space of structures is simply
$S = \R^3$ with points $(m,c,k)$. Furthermore, on physical grounds, all the parameters have to be positive, so in fact the appropriate space is
$S = \R^3_+$, which is a manifold with boundary. This representation of the linear SDOF system is actually over-parametrised, as will be shown.

Consider now, what sort of physics one might be interested in analysing. In this case, an obvious choice is the {\em free decay} of the system when it is
displaced from the origin and released. If the initial condition for the motion is $(Y_0, 0)$, the subsequent motion is described by \cite{WordenNL},

\begin{equation}
     y(t) = Y_0 e^{ - \zeta \omega_n t} \cos(\omega_d t)
\label{eq:l9_free_decay}
\end{equation}
where the {\em damping ratio} $\zeta$ is defined by,

\begin{equation}
     \zeta = \frac{ c }{ 2 \sqrt{m k} }
\label{eq:l9_def_zeta}
\end{equation}
the {\em undamped natural frequency} $\omega_n$ by,

\begin{equation}
     \omega_n = \sqrt{ \frac{k}{m} }
\label{eq:l9_def_wn}
\end{equation}
and the {\em damped natural frequency} $\omega_d$ by,

\begin{equation}
     \omega_d^2 = \omega_n^2 (1 - \zeta^2)
\label{eq:l9_def_wd}
\end{equation}

Now, the important quantities in equation (\ref{eq:l9_free_decay}) are all {\em ratios} of the physical quantities $(m,c,k)$, so transforming the point
to the scaled quantity $(m \alpha, c \alpha, k \alpha)$, where $\alpha \in \R$, produces a different equation of motion,

\begin{equation}
     m \alpha \ddot{y} + c \alpha \dot{y} + k \alpha y = 0
\label{eq:l9_sdofa}
\end{equation}
but {\em does not change the observed physics} of the free decay. This is true even if $\alpha < 0$. It would appear sensible to rule this out on physical
grounds, so one can restrict to $\alpha \in \R_+$. Now, $\R_+$ is actually a {\em group} under multiplication, where the defining properties of a
general group are as follows \cite{Birkhoff}.

A group $G$ is a set of objects $\{g\}$, with a binary operator $\circ$, such that:

\begin{enumerate}
\item $G$ is {\em closed} under the operation $\circ$ i.e.\ if $g_1, g_2 \in G$ then $g_1 \circ g_2 \in G$.

\item $G$ contains a unique {\em identity} $e$ such that $e \circ g = g \circ e = g ~~\forall g \in G$.

\item For each $g \in G$, there is an {\em inverse} $g^{-1}$ such that $g \circ g^{-1} =  g^{-1} \circ g = e$.
\end{enumerate}

Where there is no ambiguity, the $\circ$ operator will be omitted from equations, as in the standard notation of normal multiplication.

Now an {\em action} $A$ of the group $G$ on a space $X$ is defined as a map,

\begin{equation}
     A : G \times X \longrightarrow X
\label{eq:l9_def_act}
\end{equation}

The action on the parameter space, implicitly defined in equation (\ref{eq:l9_sdofa}), is a {\em left action} of the group $G= \R_+$ via $(m,c,k) \longrightarrow (m \alpha, c \alpha, k \alpha)$. Furthermore this action of the group {\em leaves the observed physics of interest unchanged}. Following the terminology of modern particle physics, one
can say that $S$ is {\em gauge invariant} under the action of the {\em gauge group} $R_+$ \cite{Hamilton}.

As mentioned earlier; as far as the physics is concerned, there is considerable redundancy in the definition of $S$, the space can be simplified accordingly. It is easy to
see that, by scaling $y$ and $t$ in equation (\ref{eq:l9_sdof}), the equation can be transformed into,

\begin{equation}
     \ddot{y} +  2 \zeta \dot{y} + y = 0
\label{eq:l9_ssdof}
\end{equation}
which one can regard as the {\em canonical representation} of an SDOF system. (The overdots now represent differentiation with respect to scaled time.) Obviously, many
systems can share the same canonical representation.

Unfortunately, this action does not amount to a simple coordinate transformation on $S$; however, it can be made into one by going to three scaling parameters
$(\alpha, \beta, \gamma)$ which are elements of the group $\R^3_+$, and defining the action of the group via $(m,c,k) \longrightarrow (m \alpha, c \beta,
k \gamma)$. Note that the $\alpha$, $\beta$ and $\gamma$ are not independent, but are related by the scaling parameters for $y$ and $t$. (In fact, the
scaling operations on $y$ and $t$ force a condition $\gamma/\beta = \beta/\alpha$ or $\alpha \gamma = \beta^2$.) With a single scaling parameter $\alpha$, the
best one can do in simplifying equation (\ref{eq:l9_sdof}), is to reduce it to,

\begin{equation}
     \ddot{y} +  2 \zeta \omega_n \dot{y} + \omega_n^2 y = 0
\label{eq:l9_sssdof}
\end{equation}

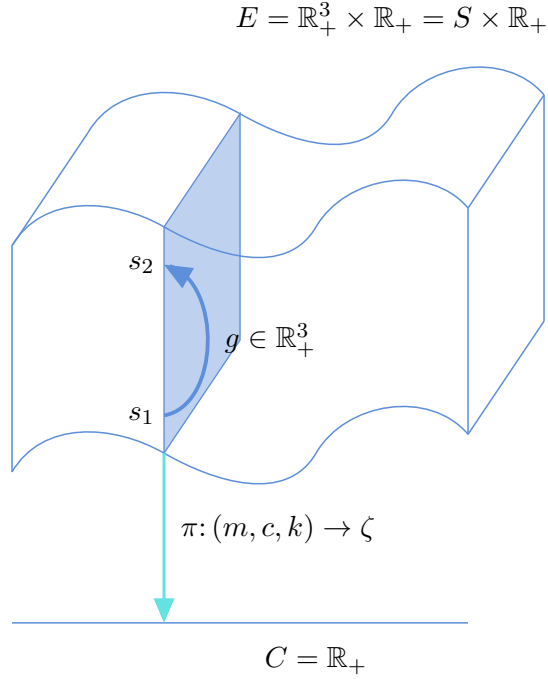
\begin{figure}[htbp!]
    \centering
     \begin{tikzpicture}
     \definecolor{blue1}{RGB}{93, 143, 218}
     \definecolor{teal}{RGB}{100, 225, 225}
     
     \draw[line width=0.2mm, blue1] (0.0, 4.0) to[out=60, in=150] (2.0, 4.25) to[out=330, in=240] (4.0, 4.25) to[out=60, in=130] (6.0, 4.5);
     \draw[line width=0.2mm, blue1] (0.0, 7.0) to[out=60, in=150] (2.0, 7.25) to[out=330, in=240] (4.0, 7.25) to[out=60, in=130] (6.0, 7.5);
     \draw[line width=0.2mm, blue1] (0.0, 4.0) to[] (0.0, 7.0);
     \draw[line width=0.2mm, blue1] (6.0, 4.5) to[] (6.0, 7.5);
     
     \draw[line width=0.2mm, blue1] (0.0, 7.0) to[] (1.0, 8.5);
     \draw[line width=0.2mm, blue1] (6.0, 7.5) to[] (7.0, 9.0);
     
     \draw[line width=0.2mm, blue1] (1.0, 8.5) to[out=60, in=150] (3.0, 8.75) to[out=330, in=240] (5.0, 8.75) to[out=60, in=130] (7.0, 9.0);
     
     \draw[line width=0.2mm, blue1] (6.0, 4.5) to[] (7.0, 6.0);
     \draw[line width=0.2mm, blue1] (7.0, 6.0) to[] (7.0, 9.0);
     
     \draw[line width=0.2mm, blue1] (0.0, 2.0) to[] (6.0, 2.0);
     
     \fill[blue1, fill opacity=0.4] (2.0, 4.25) to ++(0.0, 3.0) to ++(1.0, 1.5) to ++(0.0, -3.0) to ++(-1.0, -1.5);
     \draw[blue1, line width=0.5] (2.0, 4.25) to ++(0.0, 3.0) to ++(1.0, 1.5) to ++(0.0, -3.0) to ++(-1.0, -1.5);

     \draw[-{>[scale=1.5, length=1, width=1]}, line width=0.5mm, blue1] (2.0, 4.75) to[out=10, in=350] (2.0, 6.75);
     
     \node[] (s1) at (1.7, 4.75) {$s_{1}$};
     \node[] (s2) at (1.7, 6.75) {$s_{2}$};
     
     \node[] (g) at (3.4, 5.75) {$g \in \mathbb{R}_{+}^{3}$};
     
     \node[] (E) at (5.0, 10.0) {$E = \mathbb{R}_{+}^{3} \times \mathbb{R}_{+} = S \times \mathbb{R}_{+}$};
     
     \draw[-{>[scale=2.5, length=2, width=3]}, line width=0.4mm, color=teal] (2.0, 4.25) to (2.0, 2.0);
     
     \node[] (asd) at (3.5, 3.25) {$\pi \colon (m,c,k) \to \zeta$};
     
     \node[] (C) at (4.0, 1.5) {$C = \mathbb{R}_{+}$};
     
     \end{tikzpicture}
    \caption{Space of SDOF systems as a bundle over the space of `canonical' representations of SDOF systems. The total space $E$ is depicted as a three-dimensional manifold since visualisation of a four-dimensional manifold cannot be achieved.}
    \label{fig:l9_S_sdof}
\end{figure}

One can now think in terms of a left action of $\R^3_+$ on $S$ which leaves the physics invariant, so the gauge group is $\R^3_+$.
This leads to the first example of a fibre bundle in dynamics (Figure \ref{fig:l9_S_sdof}); although not a very complicated one, this is nonetheless interesting
as it is also the first example showing spaces of structures, and relationships between them\footnote{Note that one should not strictly call the construction
in Figure \ref{fig:l9_S_sdof} a {\em vector} bundle, because $\R_+$ is not a vector space; for all points $x$ in $\R_+$ except zero, $-x$ is not in the space.
It is perhaps only a minor abuse to talk of vector bundles because one could always work with the logarithms of the parameters, then they would be in $\R$. The
$\alpha$ etc.\ parameters could also be represented by logarithms; the group would then simply act additively.} The base space in Figure \ref{fig:l9_S_sdof} is
denoted $C$ in order to signify the space of structures in $S$ reduced to their one-parameter canonical representations.

In this example, the fibre $F = \R^3_+$ is actually the gauge group of interest; such a bundle can be referred to as a {\em principal bundle} \cite{Eguchi}.

One can also characterise the redundancy in $S$ in a more geometrical way.

Define the {\em orbit} of a point $s = (m,c,k) \in S$ as the set of all points $s'$ such that $s' = (m \alpha, c \beta, k \gamma)$ i.e.\ all the points
reachable from $s$ via the action of $G = \R^3_+$.

Now, define an {\em equivalance relation} on $S \times S$ as follows.

$\sim$ is an equivalence relation on $S \times S$ if the following are true:

\begin{enumerate}
\item {\em Identity}: $s \sim s$.

\item {\em Reflexivity}: $s_1 \sim s_2 \Longrightarrow s_2 \sim s_1$.

\item {\em Transitivity}: $s_1 \sim s_2$ and $s_2 \sim s_3$ implies $s_1 \sim s_3$.
\end{enumerate}

Now, an equivalence class $[s]$ is defined as the set of all points $s'$ such that $s' \sim s$. The set of equivalence classes is called the {\em
quotient space under $\sim$} and is denoted $S / \sim$. Under certain specific circumstances, the quotient space inherits the structure of $S$, e.g.\
the quotient space may be a manifold if the original space is. For the sake of simplicity, it will be assumed here that quotient spaces, manifolds,
groups etc.\ are appropriately well behaved.

In the space of SDOF structures $S$, one can define an equivalence relation such that $s_1 \sim s_2$ if $s_1$ is on the same orbit of the group
action of $\R^3_+$ as $s_2$. In this case, $S / \sim \cong C$, where the symbol $\cong$ denotes homeomorphism. The orbits of the group action on points
in the bundle space are the fibres, so $S / \sim$ can also be thought of as the set of fibres. The important thing is that $S / \sim$ has only a
single point for all gauge-equivalent structures, so redundancy has again been removed, and it is simpler to work with as a `space of structures'.

This construction is an example of {\em gauge-fixing}. Another way of removing the redundancy would be to work with a single representative sample for
each class of gauge-equivalent structures. Such a choice amounts to taking a section of $E$; if the section is continuous, then it will be
homeomorphic to $C$ again.

This is all interesting in terms of representing structures as points in spaces; however, it isn't immediately of use for learning in populations of
structures, which is the aim of data-based PBSHM. The next section will consider how fibre bundles can be used to represent collective feature spaces
over populations of structures.

\section{Fibre Bundles as Feature Spaces over Populations of Structures}
\label{sec:pbshm_bundles}

Suppose that one has identified the population of structures of interest and embodied that as a `space' of structures $S$. Now, further suppose that the
objective is to carry out data-based SHM on and across this space i.e.\ the ultimate aim will be to allow data-based inferences to transfer between the
structures \cite{PBSHMMSSP3}. Each member of the population $s_i$ will have an associated feature space $F_i$ (it is serendipitous that one can use the same
symbol as for `fibre'). For the moment, it will be assumed that all the feature spaces are {\em dimensionally equivalent} and {\em physically equivalent},
which means they can all be assigned the same fibre space $F$.

\begin{figure}[htbp!]
    \centering
     \begin{tikzpicture}[scale=0.87, every node/.style={scale=0.87}]
     \definecolor{blue1}{RGB}{93, 143, 218}
     \definecolor{teal}{RGB}{100, 225, 225}
     \draw[line width=0.2mm, blue1] (0.0, 0.5) to (6.0, 0.5);
     \draw[line width=0.2mm, blue1] (6.0, 0.5) to[] (7.0, 2.0);
     \draw[line width=0.2mm, blue1] (1.0, 2.0) to (7.0, 2.0);
     \draw[line width=0.2mm, blue1] (1.0, 2.0) to[] (0.0, 0.5);
     
     
     \draw[line width=0.2mm, blue1] (0.0, 4.0) to[] (0.0, 7.0);
     \draw[line width=0.2mm, blue1] (6.0, 4.5) to[] (6.0, 7.5);
     
     \draw[line width=0.2mm, blue1] (0.0, 7.0) to[] (1.0, 8.5);
     \draw[line width=0.2mm, blue1] (6.0, 7.5) to[] (7.0, 9.0);
     
     \draw[line width=0.2mm, blue1] (1.0, 8.5) to[out=60, in=150] (3.0, 8.75) to[out=330, in=240] (5.0, 8.75) to[out=60, in=130] (7.0, 9.0);
     
     \draw[line width=0.2mm, blue1] (6.0, 4.5) to[] (7.0, 6.0);
     \draw[line width=0.2mm, blue1] (7.0, 6.0) to[] (7.0, 9.0);
     
     \node[] (M) at (7.0, -0.0) {$S = \mathbb{R}_{+}^{6} = \{(l, w, t, \rho, E, \nu)\}$};
     
     \node[] (E) at (7.6, 5.0) {$E = F \times \mathbb{R}_{+}^{6}$};
     
     \node[circle,color=blue1, fill=blue1, inner sep=0pt,minimum size=3pt] (fb1) at (2.0, 4.75) {};
     \node[circle,color=blue1, fill=blue1, inner sep=0pt,minimum size=3pt] (fb2) at (2.0, 7.75) {};
     \draw[line width=0.5mm, blue1] (fb1) to[] (fb2);
     \node[] (F) at (2.2, 5.75) {$F$};
     
     \node[circle,color=black, fill=blue1, inner sep=0pt,minimum size=3pt] (x) at (2.0, 1.0) {};
     \draw[-{>[scale=2.5, length=2, width=3]}, line width=0.4mm, color=teal] (fb1) to (x);
     
     \node[] (x_) at (2.3, 1.0) {$s$};
     \node[] (x_) at (2.5, 2.25) {$\pi$};
     
     \draw[line width=0.2mm, blue1] (0.0, 4.0) to[out=60, in=150] (2.0, 4.25) to[out=330, in=240] (4.0, 4.25) to[out=60, in=130] (6.0, 4.5);
     \draw[line width=0.2mm, blue1] (0.0, 7.0) to[out=60, in=150] (2.0, 7.25) to[out=330, in=240] (4.0, 7.25) to[out=60, in=130] (6.0, 7.5);
     
     \end{tikzpicture}
    \caption{Collected feature spaces across a population of structures $S$ considered as a vector bundle.}
    \label{fig:l9_feat_bun}
\end{figure}
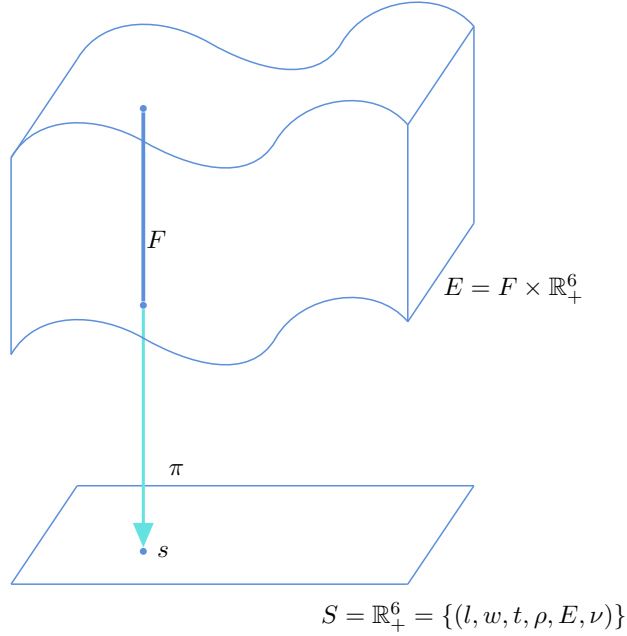

To make things specific, it will be assumed that the features in each case are the first four natural frequencies of the structure in question. In this case,
one can regard the totality of feature spaces as an $\R^4_+$-bundle over $S$. As mentioned earlier, it will not always be possible to insist that $S$
is a manifold, but this will hold for the first simple example here. Assume that $S$ is the space of cantilever beams made of homogeneous and isotropic
materials. (For the purposes of modal analysis, one can also assume linear elasticity.) In this case, the cantilever is uniquely specified by its physical
dimensions (length $l$, width $w$ and thickness $t$), and material constants (density $\rho$, Young's modulus $E$ and Poisson's ratio $\nu$), so
$S \cong \R^6_+$. So $S$ is again a (flat) manifold with boundary, and the {\em feature bundle} associated with the problem is as shown in Figure \ref{fig:l9_feat_bun}.

The first question one might ask is: why is a bundle needed, why not a single set of natural frequencies for each $s \in S$? The answer is that the problem
of interest is SHM and the features will change as damage occurs; the feature space is needed to cover the range of healthy states of $S$. Again, to make
things specific, assume that the damage can be characterised as loss of stiffness, and this can be modelled as loss of Young's modulus. One defines
$d \in [0,1]$ such that $E_d = (1-d) E$ such that $d=0$ corresponds to {\em normal condition} and $d=1$ corresponds to maximal damage. So, the fibre above
$s = (l,w,t,\rho,E,\nu)$ is the set of natural frequencies of all beams with properties $(l,w,t,\rho,E_d,\nu)$. Now, the theoretical natural frequencies are
actually the solutions to an eigenvalue problem determined by $s$, so it is not immediately clear that the fibres will glue together so that the natural
frequencies for a given state form a nice section of the bundle.

Note the implicit assumption here that the structures represented in the base manifold are {\em healthy} structures. Damage conditions are thus represented
by `state variables' in the bundle space, i.e.\ $d$ in the current example. The problem of ensuring this condition is practical or operational, rather than
geometrical, and will not be discussed further in this paper; however, it will be discussed in future publications on the PBSHM framework.

Before considering this problem a little further, it is necessary to point out another simplification that is implicitly being made here -- which is that
the features associated with (above a) given structure $s$  can be considered to be a `point' in $F$. In general, a fundamental part of any data-based SHM
strategy is the existence of training data. This requirement means that the feature space above $s$ will actually contain many points, and there will need to
be points for each damage condition that one seeks to classify (and means in turn, that there will need to be at least one index $d_i$ for the $i^{th}$
condition). For simplicity again, assume that the SHM problem is {\em novelty detection} i.e.\ one seeks only to establish
if damage is present within the structure. This is an unsupervised learning problem and can be accomplished if data are known, characteristic of the normal
(undamaged) state \cite{Farrar}. A basic outlier analysis would require enough data to construct a sample mean vector for the normal damage features, and a
sample covariance matrix. If one then regards the mean and covariance as fixed feature data characteristic of the normal condition, one can regard them as
points in an {\em associated} feature space which contains `deterministic' quantities. If one had a univariate feature space with feature $x$, the associated
feature space would be parametrised by $\overline{x}$ and $\sigma_x$. The alternative approach would be to take a fully probabilistic view
of the bundles, and it is not immediately clear how to accomplish that, although it will be the subject of future research. For the sake of simplifying the
geometry here, it will be assumed that a health state of $s$ will be represented by a single point in the fibre over $s$.

\begin{figure}[htbp!]
    \centering
    \begin{tikzpicture}[scale=0.9, every node/.style={scale=0.9}]
     \definecolor{blue1}{RGB}{93, 143, 218}
     \definecolor{teal}{RGB}{100, 225, 225}
     \definecolor{gray1}{RGB}{127, 152, 154}
     \draw[line width=0.2mm, blue1] (0.0, 0.0) to[out=30, in=150] (3.0, 0.5) to[out=330, in=210] (6.0, 1.0);
     \draw[line width=0.2mm, blue1] (6.0, 1.0) to[] (7.0, 3.0);
     \draw[line width=0.2mm, blue1] (1.0, 2.0) to[out=30, in=150] (3.0, 2.5) to[out=330, in=210] (7.0, 3.0);
     \draw[line width=0.2mm, blue1] (1.0, 2.0) to[] (0.0, 0.0);
     
     
     \draw[line width=0.2mm, blue1] (0.0, 7.0) to[out=60, in=150] (2.0, 7.25) to[out=330, in=240] (4.0, 7.25) to[out=60, in=130] (6.0, 7.5);
     \draw[line width=0.2mm, blue1] (0.0, 4.0) to[] (0.0, 7.0);
     \draw[line width=0.2mm, blue1] (6.0, 4.5) to[] (6.0, 7.5);
     
     \draw[line width=0.2mm, blue1] (0.0, 7.0) to[] (1.0, 8.5);
     \draw[line width=0.2mm, blue1] (6.0, 7.5) to[] (7.0, 9.0);
     
     \draw[line width=0.2mm, blue1] (1.0, 8.5) to[out=60, in=150] (3.0, 8.75) to[out=330, in=240] (5.0, 8.75) to[out=60, in=130] (7.0, 9.0);
     
     \draw[line width=0.2mm, blue1] (6.0, 4.5) to[] (7.0, 6.0);
     \draw[line width=0.2mm, blue1] (7.0, 6.0) to[] (7.0, 9.0);
     
     \node[] (M) at (7.0, 2.0) {$S$};
     
     \node[] (E) at (7.6, 7.0) {$E$};
     
     
     \fill[gray1, fill opacity=0.4] (0.0, 5.0) to (1.0, 6.5) to[out=60, in=150] (3.0, 6.75) to[out=330, in=240] (5.0, 6.75) to[out=60, in=130] (7.0, 7.0) to (6.0, 5.5) to[out=130, in=60] (4.0, 5.25) to[out=240, in=330] (2.0, 5.25) to[out=150, in=60] (0.0, 5.0);

    \node[circle,color=black, fill=black, inner sep=0pt,minimum size=3pt] (s10) at (2.0, 1.2) {};
    \node[] (s1) at (2.0, 1.0) {\scriptsize $s_{1}$};
    
    \node[circle,color=black, fill=black, inner sep=0pt,minimum size=3pt] (s20) at (1.8, 2.0) {};
    \node[] (s1) at (1.52, 2.0) {\scriptsize $s_{2}$};
    
    \node[circle,color=black, fill=black, inner sep=0pt,minimum size=3pt] (s30) at (2.5, 2.2) {};

    \node[circle,color=black, fill=black, inner sep=0pt,minimum size=3pt] (s40) at (3.0, 1.5) {};
    \node[] (s1) at (3.2, 1.5) {\scriptsize $s_{4}$};
    
    \node[circle,color=black, fill=black, inner sep=0pt,minimum size=3pt] (s00) at (2.3, 1.7) {};
    \node[] (s1) at (2.3, 1.5) {\scriptsize $s$};
    
    \node[circle,color=black, fill=black, inner sep=0pt,minimum size=3pt] (s50) at (5.7, 1.8) {};
    \node[] (s1) at (5.5, 1.8) {\scriptsize $s_{5}$};
    
    \node[circle,color=black, fill=black, inner sep=0pt,minimum size=3pt] (s11) at (2.0, 5.7) {};

    \node[circle,color=black, fill=black, inner sep=0pt,minimum size=3pt] (s31) at (2.5, 6.7) {};
    \node[] (s1) at (2.9, 6.7) {\tiny $n(s_{3})$};
    
    \node[circle,color=black, fill=black, inner sep=0pt,minimum size=3pt] (s41) at (3.0, 6.0) {};
    \node[] (s1) at (3.40, 6.0) {\tiny $n(s_{4})$};
    
    \node[circle,color=black, fill=black, inner sep=0pt,minimum size=3pt] (s01) at (2.3, 6.2) {};

    \node[circle,color=black, fill=black, inner sep=0pt,minimum size=3pt] (s51) at (5.7, 6.3) {};
    \node[] (s1) at (5.3, 6.3) {\tiny $n(s_{5})$};

     \draw[line width=0.4mm, color=teal] (s50) to (s51);
     \draw[line width=0.4mm, color=teal] (s40) to (s41);
     \draw[line width=0.4mm, color=teal] (s30) to (s31);
     
     \draw[line width=0.4mm, color=teal] (s10) to (s11);
     \draw[line width=0.4mm, color=teal] (s00) to (s01);
     
     \node[] (s1) at (2.3, 5.55) {\tiny $n(s_{1})$};
     \node[] (s1) at (2.35, 6.35) {\tiny $n(s)$};
     \node[] (s1) at (2.5, 2.05) {\scriptsize $s_{3}$};
     \node[circle,color=black, fill=black, inner sep=0pt,minimum size=3pt] (s21) at (1.8, 6.5) {};
     \draw[line width=0.4mm, color=teal] (s20) to (s21);
     
     \draw[line width=0.2mm, blue1] (0.0, 4.0) to[out=60, in=150] (2.0, 4.25) to[out=330, in=240] (4.0, 4.25) to[out=60, in=130] (6.0, 4.5);
     \draw[line width=0.3mm, black] (0.0, 5.0) to (1.0, 6.5) to[out=60, in=150] (3.0, 6.75) to[out=330, in=240] (5.0, 6.75) to[out=60, in=130] (7.0, 7.0) to (6.0, 5.5) to[out=130, in=60] (4.0, 5.25) to[out=240, in=330] (2.0, 5.25) to[out=150, in=60] (0.0, 5.0);

    \node[] (s1) at (1.46, 6.4) {\tiny $n(s_{2})$};
     
     \end{tikzpicture}
    \caption{Normal section $n(s)$ of a feature bundle over a space of structures $S$. It is assumed that normal condition data are not known for the point $s$.
         Interpolation/transfer of the normal condition from any neighbouring points $s_i$ should only be considered if they are sufficient close to $s$ in some
         metric on $S$. In this diagrammatic example, $s_5$ might be considered `too far away'; for transfer.}
    \label{fig:l9_norm_sec}
\end{figure}
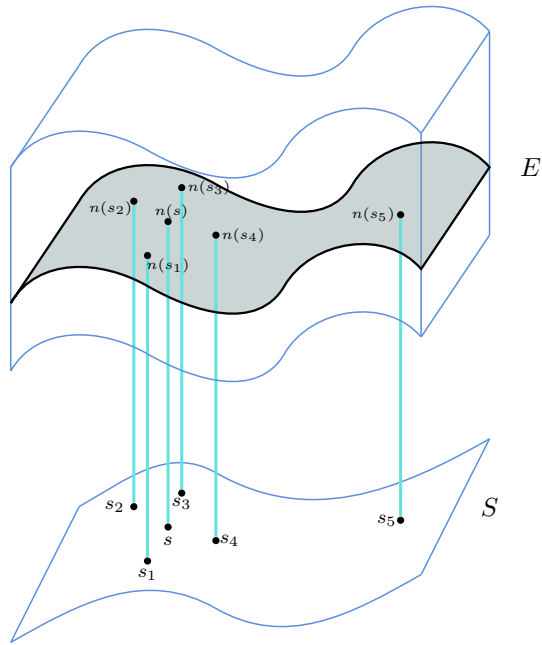

This discussion points to the existence of an important, if not critical structure in the problem. If one is to carry out novelty detection at any structure
in $s$, one needs training data or feature data which characterise the normal condition of $s$. Assuming that this can be characterised by a single point, it
is clear that the normal condition states over the whole population, determine a section of the feature bundle; this will be referred to as the {\em normal
section} for the population (Figure \ref{fig:l9_norm_sec}). Another way of characterising the basic damage detection problem boils down to the problem of specifying
the normal section across the population from normal condition data measured on only a subset of the population. In the simplest terms, one might regard this as
an interpolation problem. In the univariate case discussed above, suppose that the data $\overline{x}$ and $\sigma_x$ are known at sufficiently many training
structures, that the values can be interpolated onto neighbouring structures with no training data; in this case, the interpolated values can be used for outlier
analysis when monitored data become available at the structures without training data. Of course, the problems will generally involve multi-class classification,
and even novelty detection will need more sophisticated approaches than basic outlier analysis; in these cases, the idea will be to use transfer learning
\cite{PBSHMMSSP3}. This discussion raises an important point; while interpolation/transfer might be a powerful tool here, one must be careful not to {\em extrapolate}.
This issue means that any transfer from a structure $s$ to a `neighbouring' structure $s'$, should only be allowed if $s$ and $s'$ are {\em sufficiently
close} to each other, as measured in some appropriate metric on $S$. All of the spaces of structures discussed so far have been flat, as in fact, have all the
bundle spaces, so the standard Euclidean metric could be employed. For more complicated spaces of structures, which may not even be manifolds, it will be
necessary to establish a `metric' of some form. The most general space of structures envisioned in these foundations for PBSSHM is the complex network of attributed
graphs mentioned at various points earlier; that this space has a metric structure is demonstrated and discussed in \cite{PBSHMMSSP2}, in this series.

A further remark on interpolation and approximation in curved spaces, is that any derivatives involved in the analysis could be {\em covariant} derivatives and these
would need to be estimated from (potentially sparse) data. There is theory available for such numerical analysis in manifolds, and it is discussed briefly in the
context of SHM in \cite{Mihaylov}. Again, all the bundles discussed here so far have been flat and globally trivial.

One can now briefly return to the problem alluded to earlier, that the feature spaces may not glue together nicely to form a bundle with nice sections. Consider
the problem of mode-swapping. One normally orders the natural frequencies of a structure in order of magnitude; to simplify matters suppose that one focusses on
only two modes: a bending mode and a torsional mode. It may happen, that as one varies the structure $s$ continuously in $S$ (i.e.\ by smoothly varying $E$
in the case of the cantilever beams), the natural frequencies may cross each other i.e.\ one might go from the bending mode as lowest frequency, to the
torsional mode. If one does not track this phenomenon, and simply keeps the order of frequencies from the eigenvalue problem, the normal section will
appear to have a $C^1$ discontinuity. This exchange of modes could also occur as damage progresses in a single structure $s$ e.g.\ in the cantilever case
discussed above, as $d$ is varied. For the moment, it will be assumed that these problems do not occur; feature bundles and the sections of interest will be
assumed to be well-behaved. Even if singularities like those mentioned do not cause basic problems with the geometry, they would be likely to cause issues
with interpolation/transfer, and should be avoided if possible.

In all the cases considered so far, the overall construction of the bundles is not an issue; this is because they are all trivial bundles i.e.\ the total
spaces are all globally trivial. In fact, all bundles over {\em contractible}\footnote{A space is contractible, if it can be continuously deformed (shrunk) to a
point within the space; this is essentially trivial topology. All $\R^n_+$ are contractible, as are $\R^n$.} base spaces are trivial \cite{Eguchi}.

\section{Feature Bundles and Confounding Influences}
\label{sec:confound}

It is important to discuss how another important issue in SHM might impact on a geometrical formulation, this is the issue of {\em confounding influences}.
In order for feature data to be useful for SHM purposes, they must clearly be sensitive to damage. The natural frequencies discussed so far are candidate
features because they are damage sensitive. For example, when a crack grows in a structure, it will reduce the stiffness locally, and there may also be
friction because of rubbing of the crack interfaces; both of these mechanisms will reduce the natural frequencies. The problem is that many {\em benign}
variations in the operational conditions and/or environment may also reduce the natural frequencies e.g.\ an increase in temperature, or traffic on a bridge
will reduce frequencies. If one is using novelty detection for SHM, one is essentially only looking for changes in the features; if the features change
because of benign changes to the environment, one will potentially produce a false alarm for damage. This problem has long been recognised \cite{Farrar}, a
good review in the context of standard SHM can be found in \cite{Sohn}. In general, one needs to remove the effects of benign variations before applying the
diagnostic algorithm, this is a process often called {\em data normalisation}. In broad terms, one can divide data normalisation algorithms into
{\em subtraction} and {\em projection} schemes \cite{WordenIEEE}.

In order to discuss these matters, the earlier example of the space of cantilever beams will suffice. Suppose that some subset of the properties of the
beams are temperature dependent (in reality, they all are to some extent or other). For simplicity, it will be assumed that only the Young's modulus $E(\theta)$ is
a function of the temperature $\theta$. In practice the temperature variations will be restricted to some interval $[\theta_{min},\theta_{max}]$. If the vector of
natural frequencies is now denoted by $\uf(s)$, it is clear that they will also be functions of $\theta$. Furthermore, because this is an SHM problem, the frequencies
are also functions of $d$, the damage severity variable. So $\uf(s) = \uf(s,\theta,d)$, by virtue of the fact that $E = E(s,\theta,d)$. Now, the points on the normal
section of the feature bundle are those points corresponding to $d = 0$, but they are still functions of $\theta$. When one is measuring features for SHM, the
`physics of interest' is whether the structure is damaged or not; this means that the variations due to temperature (or any other confounding variable) can be
regarded as {\em gauge variations}. The difference here is that one cannot easily assign a group action to the variations; in the first place $\theta$
takes values on an interval, so can not be given a group structure directly (although one could monotonically transform that range onto the whole of $\R$);
secondly, the effect of temperature on the natural frequencies is complicated, so the action will not generally allow an analytical formulation. Putting
aside these issues, one can still regard confounding variables as gauge degrees of freedom. Apart from the simplicity of removing gauge variables, they
are actually a nuisance in this problem. It is clear that a form of gauge-fixing is needed, and this is essentially what data normalisation is. Note that, like the
damage state variables, the different temperature states available to a given structure are represented in the feature bundle; the representation of the structure
in the base space will correspond to a reference temperature. There is no restriction to a common reference temperature in the space of structures; variations can
be accommodated as long as appropriate data normalisation is applied. (This freedom will be clearly illustrated in the case study in the second part of this paper.)

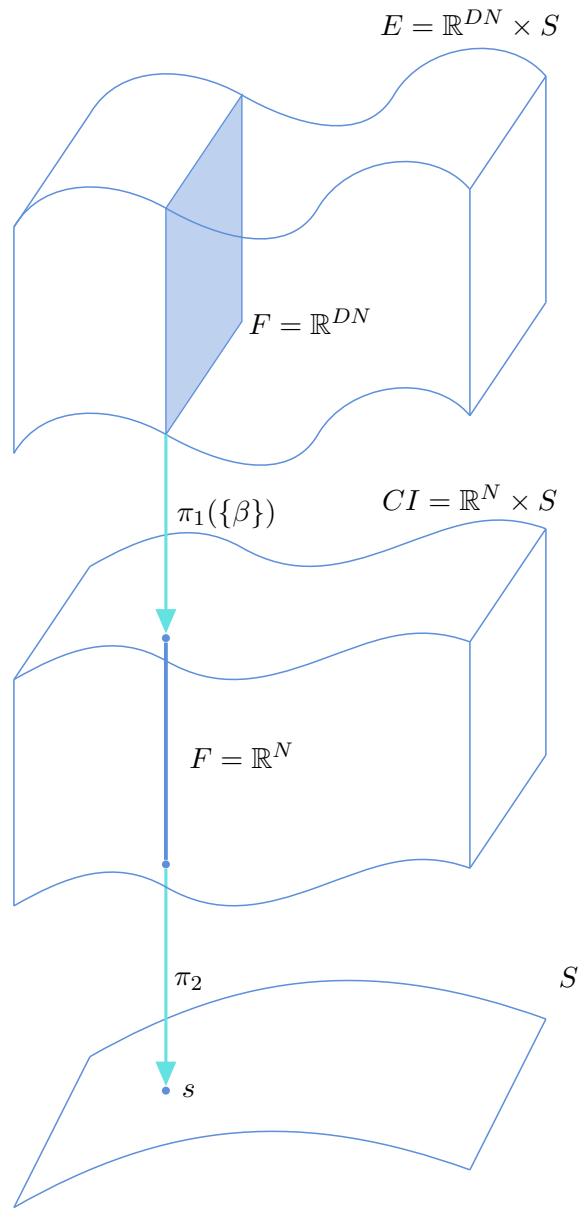
\begin{figure}[htbp!]
    \centering
     \begin{tikzpicture}
     \definecolor{blue1}{RGB}{93, 143, 218}
     \definecolor{teal}{RGB}{100, 225, 225}
     
     \draw[line width=0.2mm, blue1] (0.0, 4.0) to[out=60, in=150] (2.0, 4.25) to[out=330, in=240] (4.0, 4.25) to[out=60, in=130] (6.0, 4.5);
     \draw[line width=0.2mm, blue1] (0.0, 7.0) to[out=60, in=150] (2.0, 7.25) to[out=330, in=240] (4.0, 7.25) to[out=60, in=130] (6.0, 7.5);
     \draw[line width=0.2mm, blue1] (0.0, 4.0) to[] (0.0, 7.0);
     \draw[line width=0.2mm, blue1] (6.0, 4.5) to[] (6.0, 7.5);
     
     \draw[line width=0.2mm, blue1] (0.0, 7.0) to[] (1.0, 8.5);
     \draw[line width=0.2mm, blue1] (6.0, 7.5) to[] (7.0, 9.0);
     
     \draw[line width=0.2mm, blue1] (1.0, 8.5) to[out=60, in=150] (3.0, 8.75) to[out=330, in=240] (5.0, 8.75) to[out=60, in=130] (7.0, 9.0);
     
     \draw[line width=0.2mm, blue1] (6.0, 4.5) to[] (7.0, 6.0);
     \draw[line width=0.2mm, blue1] (7.0, 6.0) to[] (7.0, 9.0);
     
     \node[] (E) at (6.0, 9.7) {$E=\mathbb{R}^{DN} \times S$};
     
     
     \draw[line width=0.2mm, blue1] (0.0, 1.0) to[out=30, in=150] (2.0, 1.25) to[out=330, in=200] (4.0, 1.25) to[out=20, in=160] (6.0, 1.5);
     \draw[line width=0.2mm, blue1] (0.0, -2.0) to[] (0.0, 1.0);
     \draw[line width=0.2mm, blue1] (6.0, -1.5) to[] (6.0, 1.5);
     
     \draw[line width=0.2mm, blue1] (0.0, 1.0) to[] (1.0, 2.5);
     \draw[line width=0.2mm, blue1] (6.0, 1.5) to[] (7.0, 3.0);
     
     \draw[line width=0.2mm, blue1] (1.0, 2.5) to[out=30, in=150] (3.0, 2.75) to[out=330, in=200] (5.0, 2.75) to[out=20, in=160] (7.0, 3.0);
     
     \draw[line width=0.2mm, blue1] (6.0, -1.5) to[] (7.0, 0.0);
     \draw[line width=0.2mm, blue1] (7.0, 0.0) to[] (7.0, 3.0);
     
     \node[] (E) at (6.0, 3.4) {$CI=\mathbb{R}^{N} \times S$};
     
     \node[] (F) at (3.9, 5.75) {$F=\mathbb{R}^{DN}$};
     
     \fill[blue1, fill opacity=0.4] (2.0, 4.25) to ++(0.0, 3.0) to ++(1.0, 1.5) to ++(0.0, -3.0) to ++(-1.0, -1.5);
     \draw[blue1, line width=0.5] (2.0, 4.25) to ++(0.0, 3.0) to ++(1.0, 1.5) to ++(0.0, -3.0) to ++(-1.0, -1.5);
     
     \node[circle,color=black, fill=blue1, inner sep=0pt,minimum size=3pt] (f2a) at (2.0, 1.55) {};
     \draw[-{>[scale=2.5, length=2, width=3]}, line width=0.4mm, color=teal] (2.0, 4.25) to (f2a);

     \node[circle,color=black, fill=blue1, inner sep=0pt,minimum size=3pt] (f2b) at (2.0, -1.45) {};
     \draw[line width=0.5mm, blue1]  (f2a) to (f2b);
     
     \node[] (pi1) at (2.8, 3.2) {$\pi_{1}(\{\beta\})$};
     \node[] (F2) at (3.0, 0.0) {$F=\mathbb{R}^{N}$};
     
     \draw[line width=0.2mm, blue1] (0.0, -6.0) to[out=30, in=160] (6.0, -5.5);
     \draw[line width=0.2mm, blue1] (1.0, -4.0) to[out=30, in=160] (7.0, -3.5);
     \draw[line width=0.2mm, blue1] (0.0, -6.0) to (1.0, -4.0);
     \draw[line width=0.2mm, blue1] (6.0, -5.5) to (7.0, -3.5);
     
     \node[circle,color=black, fill=blue1, inner sep=0pt,minimum size=3pt] (spoint) at (2.0, -4.45) {};
     \draw[-{>[scale=2.5, length=2, width=3]}, line width=0.4mm, color=teal] (f2b) to (spoint);
     
     \node[] (pi2) at (2.3, -3.0) {$\pi_{2}$};
     
     \node[] (ss) at (2.3, -4.45) {$s$};
     
     \node[] (sss) at (7.3, -2.95) {$S$};
     
     \draw[line width=0.2mm, blue1] (0.0, -2.0) to[out=30, in=150] (2.0, -1.75) to[out=330, in=200] (4.0, -1.75) to[out=20, in=160] (6.0, -1.5);
     
     \end{tikzpicture}
    \caption{Sequence of bundles arising from cointegration of time series features.}
    \label{fig:l9_coint_bundle}
\end{figure}

One approach to the data normalisation problem in this context, is essentially a subtraction scheme; it assumes that measurements of the temperature are
available, so that points $\uf(s_i,\theta,0)$ are known on some subset of structures $s_i$; if this is the case, one can fit a series of regression models
like the linear,

\begin{equation}
     \uf(s_i,\theta,0) = \uf(s_i,0,0) + \uf_1 \theta + ...
\label{eq:l9_reg}
\end{equation}
where $\uf_1$ is a vector of regression coefficients. This allows one to gauge fix to the normal condition features $\uf(s_i,0,0)$, and these can, in
principle, be transferred onto neighbouring members of the populations -- systems $s$ -- which do not have training data.

Projection methods offer a more geometrical solution to the problem; this also allows for the discussion of time-series features, which have not appeared up
to now. Suppose that the feature data for a given system $s_i$ are multivariate time series, which have been sampled over a time $[0,T_i]$ and have $N_i$
samples. The features are thus represented by an $N_i \times D_i$ matrix $[X]$ where $D_i$ is the number of time series variables. This is too general
for immediate use, so it will be assumed that $T_i = T$, $N_i = N$ and $D_i = D$, for fixed $T$, $N$ and $D$. This is quite a high-dimensional feature space;
it is $\R^{DN}$, where $N$ might be large\footnote{As before, the fact that there will generally be multiple samples of training data is ignored; it is assumed
that each structure has a `point' in the feature space corresponding to each damage state of interest.}. In general, the quantities measured in structural dynamics:
displacements,velocities, accelerations etc.\ will be zero-mean as stochastic processes, so the feature spaces and thus fibres in the feature bundle are vector spaces
in this case.

One approach to projection -- and arguably the state of the art -- is {\em cointegration} \cite{Cross1,Cross2}. In this approach, the confounding influences are
considered to be common nonstationary trends in the data which can be removed by forming appropriate combinations of the components of the multivariate
series. The details of the algorithms etc.\ are not relevant here -- the curious reader can consult the original papers \cite{Cross1,Cross2}. Other
benefits of cointegration, and projection methods in general \cite{Manson,Kullaa}, are that one does not require measurements of the confounding variables, and that
multiple influences can be removed in one step. In the case under discussion here, cointegration, via a linear combination with coefficients $\underline{\beta}$,
reduces the $D$ nonstationary time series -- the columns of $[X]$ -- to a single stationary time series (arrayed in a vector) $\ux$, which has been purged of its
temperature variation i.e.\ has been gauge-fixed. The geometry of the situation is shown in Figure \ref{fig:l9_coint_bundle}.

In this geometrical context, cointegration is a {\em bundle map}. It is interesting, and may be important to note that, as cointegration produces a zero-mean
residual time series, the normal section in $CI$ (see Figure \ref{fig:l9_coint_bundle}), is actually the {\em zero section} of the bundle \cite{Eguchi}. In
terms of interpolation/transfer, the cointegration vectors can potentially be transferred between members of the population, from a subset with training data,
to those with none.

It is worth considering geometrical issues which might arise in dealing with cointegrating vectors. Returning to the examples of the cantilever beams; suppose one
considers the features that might arise from the situation where $\theta$ and $d$ both vary. The possible features will live on a two-dimensional submanifold
of the four-dimensional fibre (four natural frequencies); however, both temperature increase and damage can cause the natural frequencies to decrease. This
observation means that a feature might arise from temperature decrease alone or damage alone, so the `submanifold' will actually self-intersect, so can not
actually be a manifold. This observation will bear further investigation.

\section{Feature Bundles and More Complicated Spaces of Structures}
\label{sec:complicated}

In order to motivate the discussion of this section, it will consider the {\em Irreducible Element} (IE) and {\em Attributed Graph} (AG) representations of
structures as discussed in \cite{PBSHMMSSP2,PBSHMMSSP3}.

If the population of structures of interest is {\em homogeneous} \cite{PBSHMMSSP1}, each structure will be parameterised by the same number of
continuous parameters, and all corresponding parameters will have the same physical interpretation. In this case, the situation is like the space/population
of cantilever beams discussed earlier and the space of structures $S$ will be a manifold. However, in general, one might have to deal with a {\em heterogeneous
population}. As a simplified example of this situation, one might consider a population of {\em laminated cantilever beams}. Adding the number of layers $n_L$
as an explicit parameter yields a set of $6 n_L + 1$ parameters per beam $\{n_L,\{l_i,w_i,t_i,\rho_i,E_i,\nu_i\}, i=1,\ldots,n_l\}$, assuming perfect
interfaces. So the cantilevers do not look too strange, one can assume common values of $l_i$ and $w_i$. The problem is that the set of structures is not a
manifold; it is not even a topological space, because it has different dimensions in different places.

The obvious mathematical solution to the problem is to partition the set into subsets with common numbers of layers, and then one simply has multiple versions
of the original cantilever problem; more parameters, but no new geometry. Unfortunately, the whole point of the exercise is to solve problems in data-based SHM;
to move inferences from structures where one has training data, to structures where one does not. One can only partition the population, if each subset has
enough exemplars with data for transfer to be feasible within the subset/sub-population. PBSHM is proposed in the first place to deal with the problem that
data across populations may well be scarce. If pragmatism demands that one has to deal with heterogeneous populations, one has to adopt the methods appropriate
to those problems i.e.\ to match structures and transfer inferences across more complicated populations \cite{PBSHMMSSP3}.

For now, it will be assumed that there is a characterisation of the population in terms of AG representations of the structures. The most general structure one
can adopt in this situation is to assume a complex network with nodes comprised of the AGs \cite{PBSHMMSSP2}. Transfer between structures in such a population
is the detailed subject of other papers in this series \cite{PBSHMMSSP3}; the discussion here will concentrate on the geometry of the feature spaces associated with the
structures. Each AG model within the population will have varying numbers of parameters, so the space $S$ can not be a manifold. As discussed earlier, the least
one can ask of this $S$ is that it be equipped with a metric of some kind so that the `closeness' of structures can be measured, and transfer is only attempted
for structures that are appropriately close in the metric. Classical graph matching metrics are available \cite{Bunke,Fernandez}, and metrics for matching attributed
graphs have been proposed based on machine learning \cite{Li}.

\begin{figure}[htbp!]
    \centering
     \begin{tikzpicture}
     \definecolor{blue1}{RGB}{93, 143, 218}
     \definecolor{teal}{RGB}{100, 225, 225}
     
     
     \draw[line width=0.2mm, blue1] (0.0, 7.0) to[out=60, in=150] (2.0, 7.25) to[out=330, in=240] (4.0, 7.25) to[out=60, in=130] (6.0, 7.5);
     \draw[line width=0.2mm, blue1] (0.0, 4.0) to[] (0.0, 7.0);
     \draw[line width=0.2mm, blue1] (6.0, 4.5) to[] (6.0, 7.5);
     
     \draw[line width=0.2mm, blue1] (0.0, 7.0) to[] (1.0, 8.5);
     \draw[line width=0.2mm, blue1] (6.0, 7.5) to[] (7.0, 9.0);
     
     \draw[line width=0.2mm, blue1] (1.0, 8.5) to[out=60, in=150] (3.0, 8.75) to[out=330, in=240] (5.0, 8.75) to[out=60, in=130] (7.0, 9.0);
     
     \draw[line width=0.2mm, blue1] (6.0, 4.5) to[] (7.0, 6.0);
     \draw[line width=0.2mm, blue1] (7.0, 6.0) to[] (7.0, 9.0);
     
     \node[] (E) at (3.6, 9.5) {$E=F \times S$};
     
     \node[] (F) at (2.2, 5.75) {$F$};
     \node[] (F) at (3.8, 5.75) {$F$};
     
     \node[circle,color=black, fill=blue1, inner sep=0pt,minimum size=5pt] (1) at (2.5, 1.0) {};
     \node[circle,color=black, fill=blue1, inner sep=0pt,minimum size=5pt] (2) at (3.5, 1.2) {};
     \node[circle,color=black, fill=blue1, inner sep=0pt,minimum size=5pt] (3) at (3.0, 0.5) {};
     \node[circle,color=black, fill=blue1, inner sep=0pt,minimum size=5pt] (4) at (2.8, -0.5) {};
     \node[circle,color=black, fill=blue1, inner sep=0pt,minimum size=5pt] (5) at (4.8, 0.5) {};
     \node[circle,color=black, fill=blue1, inner sep=0pt,minimum size=5pt] (6) at (5.0, 1.5) {};
     \node[circle,color=black, fill=blue1, inner sep=0pt,minimum size=5pt] (7) at (1.0, 1.0) {};
     \node[circle,color=black, fill=blue1, inner sep=0pt,minimum size=5pt] (8) at (1.5, 0.2) {};
     
     \draw[line width=0.4mm, blue1] (1) to[] (2);
     \draw[line width=0.4mm, blue1] (2) to[] (3);
     \draw[line width=0.4mm, blue1] (1) to[] (3);
     \draw[line width=0.4mm, blue1] (3) to[] (4);
     \draw[line width=0.4mm, blue1] (4) to[] (5);
     \draw[line width=0.4mm, blue1] (3) to[] (5);
     \draw[line width=0.4mm, blue1] (5) to[] (6);
     \draw[line width=0.4mm, blue1] (2) to[] (6);
     \draw[line width=0.4mm, blue1] (1) to[] (7);
     \draw[line width=0.4mm, blue1] (8) to[] (7);
     \draw[line width=0.4mm, blue1] (1) to[] (8);
     \draw[line width=0.4mm, blue1] (4) to[] (8);
    
     \node[circle,color=blue1, fill=blue1, inner sep=0pt,minimum size=3pt] (fb11) at (2.5, 4.5) {};
     \node[circle,color=blue1, fill=blue1, inner sep=0pt,minimum size=3pt] (fb12) at (2.5, 7.5) {};
     \draw[line width=0.5mm, blue1] (fb11) to[] (fb12);
     
     \node[circle,color=blue1, fill=blue1, inner sep=0pt,minimum size=3pt] (fb21) at (3.5, 4.38) {};
     \node[circle,color=blue1, fill=blue1, inner sep=0pt,minimum size=3pt] (fb22) at (3.5, 7.38) {};
     \draw[line width=0.5mm, blue1] (fb21) to[] (fb22);
     
     \node[circle,color=blue1, fill=blue1, inner sep=0pt,minimum size=3pt] (fb31) at (5.0, 5.35) {};
     \node[circle,color=blue1, fill=blue1, inner sep=0pt,minimum size=3pt] (fb32) at (5.0, 8.35) {};
     \draw[line width=0.5mm, blue1] (fb31) to[] (fb32);
    
     \draw[-{>[scale=2.5, length=2, width=3]}, line width=0.4mm, color=teal] (1) to (fb11);
     
     \draw[-{>[scale=2.5, length=2, width=3]}, line width=0.4mm, color=teal] (2) to (fb21);
     
     \draw[-{>[scale=2.5, length=2, width=3]}, line width=0.4mm, color=teal] (6) to (fb31);

     
     \node[] (S) at (4.4, -0.8) {$S=CN\{AG_{i}\}$};
     
     \node[] (ag1) at (3.8, 1.6) {$AG_{1}$};
     \node[] (ag2) at (5.4, 1.7) {$AG_{2}$};
     \node[] (ag3) at (2.5, 1.4) {$AG(s)$};
     
     \draw[-{>[scale=2.5, length=2, width=3]}, line width=0.4mm, color=blue1] (2.6, 5.75) to[out=20, in=160] (3.4, 5.75);
     
     \draw[line width=0.2mm, blue1] (0.0, 4.0) to[out=60, in=150] (2.0, 4.25) to[out=330, in=240] (4.0, 4.25) to[out=60, in=130] (6.0, 4.5);
     
     \end{tikzpicture}
    \caption{Feature bundle over a more complicated space of structures. $S=CN\{AG_i\}$ is a complex network of attributed graphs. The space is equipped with a
         metric and transfer of SHM inferences between two structures would be enabled if they were sufficiently close in the metric. In the current
         schematic $AG_1$ would be considered close enough to $AG(s)$, but $AG_2$ would not.}
    \label{fig:l9_AG_bundle}
\end{figure}
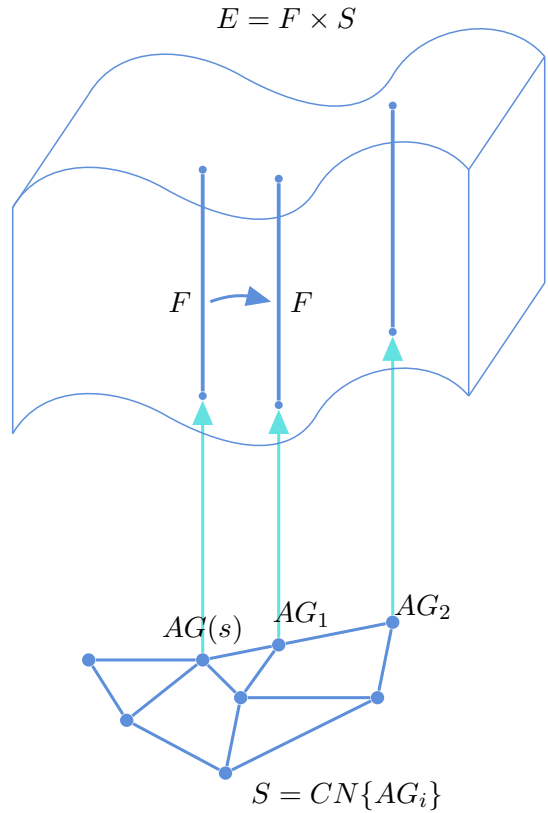

Even if the population is heterogeneous, it may be that their feature spaces are homeomorphic e.g.\ it may have been decided to monitor the first $n_f$
natural frequencies across the population. In this case, it may be possible to assemble the feature spaces into a `vector bundle' over $S$ (Figure
\ref{fig:l9_AG_bundle}). Transfer will be enabled again between fibres if the source task and target task are sufficiently close in the metric of $S$. If a
number of structures with training data are sufficiently close, multi-source transfer might be enabled \cite{Sun}.

\section{Machine Learning for Geometrical Problems with Graphs}
\label{sec:ml_geom}

While the geometrical reasoning so far has motivated some interesting observations about population-based SHM, it has not produced any clear suggestions for {\em solutions}
to the problems highlighted. In order to make progress for now, it will be necessary to apply some ideas from machine learning. For clarity, it is useful to concentrate
on one specific problem, and the remainder of this paper will be concerned with the estimation of the {\em normal condition section} over a population of structures, as
discussed in Section \ref{sec:pbshm_bundles}. Recall that the normal section is a map into the feature bundle over a population, which selects features corresponding to
the undamaged condition of a given structure represented by a point in the base space. In this case, the base space is the complex network of attributed graphs
$CN\{AG\}$. The desired map is thus of the form,

\begin{equation}
     n: CN\{AG\} \longrightarrow F
\label{eq:def_N}
\end{equation}
where $F$ is the feature space of interest, and so also the fibre of the feature bundle. In the remainder of this paper, the feature of interest will be the first natural
frequency of structure $\omega$, so the desired map is,

\begin{equation}
     n(AG_i) = \omega_i
\end{equation}
where $i$ indexes the structure/graph of interest.

As discussed, the problem is to find $n$ from a finite (probably small) training set of known structures with known/measured (first) natural frequencies:
$\{ (AG_i, \omega_i); i=1,\ldots,N_t \}$. In purely geometrical terms, this is an interpolation problem; however, it is clearly addressable using machine learning
(particularly when noting that the measurements will generally be noisy), where it would be regarded as a regression problem. The issue is that the input space is a set
of graphs, as distinct from points in some $\R^n$, which would be usual for a regression problem. This is an important distinction, which would rule out most standard
regression models like {\em artificial neural networks} (ANNs) \cite{Bishop,Bishop2}.

Because graphs are ubiquitous across many problems, learning from them has been studied a great deal, particularly in the modern context of deep learning; a useful recent
survey is \cite{Bacciu}. One way to deal with graphs, is to {\em embed} them in a more easily-understood space; this is the {\em graph embedding} problem i.e.\ one
establishes a map,

\begin{equation}
     \psi: CN\{AG\} \longrightarrow \R^m
\label{eq:defenb}
\end{equation}
where $m$ is the {\em embedding dimension}. After applying $\psi$, standard machine learning is used. A good recent survey on graph embedding is \cite{Cai}.

As one might imagine, graph embedding is not trivial, particularly if the objective is to embed in a low-dimensional space but to retain all of the important graph
structure. There are two main problems:

\begin{enumerate}
\item Graphs will generally have different numbers of nodes and edges, so it is not obvious how to embed in a space of fixed dimension $m$.

\item Graphs do not have unique representations themselves. The way that the nodes and edges are labelled is unimportant; if one permutes all the indices of the nodes,
the graph structure itself is unchanged. In mathematical terms, suppose that an operation $\pi$ permutes the node indices of a graph $G$ (and of course the edge indices,
accordingly), then the embedding should satisfy $\psi( \pi(G) ) = \psi(G)$. In an attributed graph, the operation $\pi$ would also carry the attribute vectors onto the
correct nodes and edges.
\end{enumerate}

In terms of the latter problem, one could try and learn a many-to-one mapping to project out the permutations, but this makes the learning problem more difficult. The
alternative is to try and learn $n$ and $\psi$ together, using algorithms tailored to graph inputs. This is the approach that will be followed in this paper; specifically
using {\em graph networks} \cite{Battaglia}.

Before proceeding, it will be necessary to establish some notation for graphs; where possible, this will follow \cite{PBSHMMSSP2}.

A graph $G$ will consist of a set of nodes/vertices $V$, together with a set of edges $E$, connecting the nodes; thus $G = \{V,E\}$. The number of nodes will be denoted
by $n_V$ and the number of edges by $n_E \le n_V^2$. $V = \{V_i; i=1,\ldots,n_v\}$ and $E = \{E_{ij}; 1 \le i,j, \le n_V\}$.

Which edges in a graph are actually present will be indicated by an {\em adjacency matrix} $A$, where $A_{ij} = 1$ if there is an edge between $V_i$ and $V_j$, and
$A_{ij} = 0$ otherwise. The diagonal elements $A_{ii}$, will only be non-zero if self-edges are allowed in the graph. {\em Multi-graphs} arise if a node pair is allowed
to have more than one connecting edge; this is indicated in the adjacency graph, by setting $A_{ij} = k$ if the vertices $V_i$ and $V_j$ are connected by $k$ edges. For
directed graphs, the elements $A_{ij}$ and $A_{ji}$ are used to distinguish the directions of the edges. If a permutation operator $\pi$ acts on $V$, the action of $\pi$
on $E$ is to permute the rows and columns of $A$ accordingly.

In  \cite{PBSHMMSSP2}, the attribute vectors for a node $V_i$ are denoted by $\uth_i$, and for an edge $E_{ij}$ by $\uth_{ij}$; however, in this paper it will be convenient
to adopt a notation consistent with \cite{Battaglia}. In this notation, the attribute vectors associated with a node $V_i$ will be denoted $\uv_i$. Furthermore, the edges
will be indexed $E_i$, with $i = 1,\ldots,n_E$ and their attribute vectors denoted by $\ue_i$.

In the following, graphs will also be allowed {\em global} attributes associated with the entire graph; these will be denoted $\uu$.

Suppose that an edge $E_k$ in a graph is directed between nodes $V_i$ and $V_j$, then $V_i$ is termed a {\em sender node} for the edge, and is denoted $S_k$ (with index
$s_k$); $V_j$ is termed a receiver node, and denoted $R_k$ (with index $r_k$). Unless the graph is a multi-graph, edges will only have one receiver and sender node.

With these conventions, a general attributed graph is characterised as,

\begin{equation}
     AG = \{ V, E, \{\uv\}, \{\ue\}, \uu \}
\label{eq:defag}
\end{equation}
and the machine learning problem of interest here is to learn the map $n: AG \longrightarrow \omega$, given training data $\{ (AG_i, \omega_i);~i=1,\ldots,n_T \}$. In order
to accomplish this, the map will need to be represented in terms of tunable parameters. In general, one might also include tunable parameters in the graph structures to
facilitate the learning process. In the problem considered here, the parameter $\omega$ will actually be included as a tunable global attribute. As usual, training will be
accomplished via an iterative algorithm, as specified in \cite{Battaglia}

\section{Graph Neural Networks}
\label{sec:gnns}

\subsection{Introduction}

Machine learning is a means of constructing connections between quantities of interest, based on observations of those quantities. Most of the effort so far has been
concerned with learning input-output mappings for regression or classification problems, where the input and output quantities are vectors in multidimensional real
vector spaces; many of the `classical' algorithms e.g.\ Gaussian processes, neural networks (NN) \cite{Bishop,Bishop2}, support vector machines (SVM) \cite{Cortes}, operate
in precisely this manner. In more recent years, this viewpoint has been recognised as quite restrictive in a wide range of disciplines including: social networking
\cite{Perozzi,Kipf,Zhang}, biology \cite{Dobson}, chemistry \cite{Ralaivola}, medicine \cite{Zitnik}, engineering \cite{Wang} etc. The issue has been that interest has
been focussed on relationships between objects which exist in more diverse structures than vector spaces.

A fairly trivial problem presented in \cite{Battaglia}, illustrates where such algorithms can fail. The problem is to calculate the centre of mass of a planetary
system. Assuming training data for a fixed number of planets and their resulting centroid, a `traditional' ML algorithm will probably be able to calculate the centroids
for solar systems in testing data sets constructed in the same manner. However, the algorithm would fail if one were to take as an input, a permutation of the input
vector (with masses moved as the planet indices move); it will only work if the coordinates of the planets are given in a specific order and according to the
distribution of the training data. Furthermore, if the number of planets was not fixed, it would be impossible to get an estimate of their centre of mass using the
specific model; a new model would have to be trained with appropriate data. These are the same problems highlighted earlier in terms of learning on graphs: the graphs do
not have fixed `dimension' and have a natural permutation invariance which is not respected by classical methods.

In order to solve problems like the one described above, various types of graph networks (GNs) have been developed. A large class of GNs is encompassed by the approach
introduced in \cite{Battaglia}, so that model will form the basis of the discussion here. The algorithm uses graphs as both inputs and outputs. The graphs have attributes
assigned to their nodes, their edges and also {\em globally}. With this representation of data, the algorithm becomes invariant to permutations on the inputs and also
becomes, via the proposed computational blocks, invariant to the size of the graph; i.e.\ the same model may be applied to graphs of any size. Such an algorithm is
well-suited to perform inference amongst structures, since they may differ a lot in size, materials, functionality etc. Even the layout of substructures within the greater structure plays a crucial role in the behaviour of the system \cite{PBSHMMSSP2}. The prior information provided by the user as an IE-model or AG of the
structure contains a great deal of guidance in terms of how variable and complex structures can be. The formulation is extremely versatile and existing neural network
algorithms may be considered as a subclass of the newly proposed algorithm \cite{Battaglia}.

Furthermore, by using graphs as the objects of interest, the algorithm allows different \textit{inductive biases} in the model to those encountered in classical
machine learning. An example of an inductive bias is the introduction of $L_2$ regularisation for training of a neural network; this reflects a belief that smooth
mappings should be favoured, and this can be imposed by favouring solutions with lower weight values on network connections. In classification or regression problems,
this inductive bias will select a more smooth decision boundary or approximating function from the neural network. Another widely-used inductive bias is applied in the
architecture of convolutional neural networks \cite{Geron}. By applying local convolution operators in the earlier layers, the algorithm looks for {\em local} patterns;
objects within an image for example. The intensity of each local object is extracted and used as a feature in a later fully-connected multilayer perceptron.

In the case of graphs, inductive biases are inserted into the model via connections defined in the representation of data. Such biases are well known in the field of
probabilistic graphicals models (PGMs). A good example of this is the PGM for a Markov process, where the state node at time $t$ is only connected to the state node at
time $t-1$, because it is dependent on that state and no previous ones. Different types of connections may also be motivated by inductive biases, expanding the connectivity
and functionality of the network. In the framework of PBSHM, the geometrical formulation discussed in the first part of the paper is anticipated to motivate new inductive
biases which will further empower algorithms.

The rest of this paper will be taken up with demonstrating the potential power of graph networks in addressing geometrical problems in population-based SHM, of the sort
described in the first part of the paper. Because GNN methods have not previously been used in SHM, the basic terminology and the form of the algorithm will be discussed
in some detail.

\subsection{Graph network computational blocks}

\begin{figure}[htbp!]
    \centering
    \begin{subfigure}[b]{0.49\textwidth}
        \centering
        \scalebox{0.85}{
        \begin{tikzpicture}
            
\node[circle,draw, minimum size=1cm, line width=0.5mm] (A) at  (0, 0) {};
\node[circle,draw, minimum size=1cm, line width=0.5mm] (B) at  (1, 2)  {};
\draw[-{Latex[width=5mm, length=2mm]}, line width=0.5mm] (A) to[out=40, in=270] (B);
\draw[-{Latex[width=5mm, length=2mm]}, line width=0.5mm] (B) to[out=200, in=100] (A);

\node[circle, draw, minimum size=1cm, line width=0.5mm] (C) at  (3, -0.3) {};
\node[circle, draw, minimum size=1cm, line width=0.5mm] (D) at  (2.5, 3) {\LARGE $\bm{{\underline{v}}_{i}}$};
\node[circle, draw, minimum size=1cm, line width=0.5mm] (E) at  (4.5, 1.5) {};

\draw[-{Latex[width=5mm, length=2mm]}, line width=0.5mm] (C) to (A);
\draw[-{Latex[width=5mm, length=2mm]}, line width=0.5mm] (C) to (E);
\draw[-{Latex[width=5mm, length=2mm]}, line width=0.5mm] (C) to (B);
\draw[-{Latex[width=5mm, length=2mm]}, line width=0.5mm] (D) to (E);

\draw[-{Latex[width=5mm, length=2mm]}, line width=0.5mm] (B) to[out=10, in=240] (D);
\draw[-{Latex[width=5mm, length=2mm]}, line width=0.5mm] (D) to[out=170, in=80] node[above] {\LARGE $\bm{{\underline{e}}_{k}}$} (B);

\draw[line width=0.5mm,-{Latex[width=5mm, length=2mm]},shorten >=1pt] (E) to [out=270,in=0,loop,looseness=3.4] (E);

\draw[line width=0.5mm] (4, 3.4) -- (4.7, 2.7) -- (5.4, 3.4) -- (4.7, 4.1) -- (4, 3.4);
\node[] at (4.7, 3.4) {\LARGE $\bm{{\underline{u}}}$};

\end{tikzpicture}
        }
    \end{subfigure}
    \begin{subfigure}[b]{0.49\textwidth}
        \centering
        \scalebox{0.85}{
        \begin{tikzpicture}
            
\draw[line width=0.5mm] (1.8, 4) -- (2.5, 3.3) -- (3.2, 4) -- (2.5, 4.7) -- (1.8, 4);
\node[] at (2.5, 4) {\LARGE $\bm{{\underline{u}}}$};

\node[circle, draw, minimum size=1.2cm, line width=0.5mm] (D) at  (2.5, 2.3) {\LARGE $\bm{{\underline{v}}_{i}}$};

\node[circle, dashed, draw, minimum size=1cm, line width=0.5mm] (sender) at  (1.2, 0.6) {\LARGE $\bm{{\underline{v}}_{sk}}$};

\node[circle, dashed, draw, minimum size=1cm, line width=0.5mm] (receiver) at  (3.8, 0.6) {\LARGE $\bm{{\underline{v}}_{rk}}$};

\draw[-{Latex[width=3mm, length=2mm]}, line width=0.5mm] (sender) to node[above] {\LARGE $\bm{{\underline{e}}_{k}}$} (receiver);

\node[] at (6.2, 4) {\small $\{Global\ attributes\}$};
\node[] at (6.2, 2.3) {\small $\{Node\ attributes\}$};
\node[] at (6.2, 0.6) {\small $\{Edge\ attributes\}$};

\end{tikzpicture}
        }
    \end{subfigure}
    \caption{General graph architecture.}
    \label{fig:general_graph}
\end{figure}
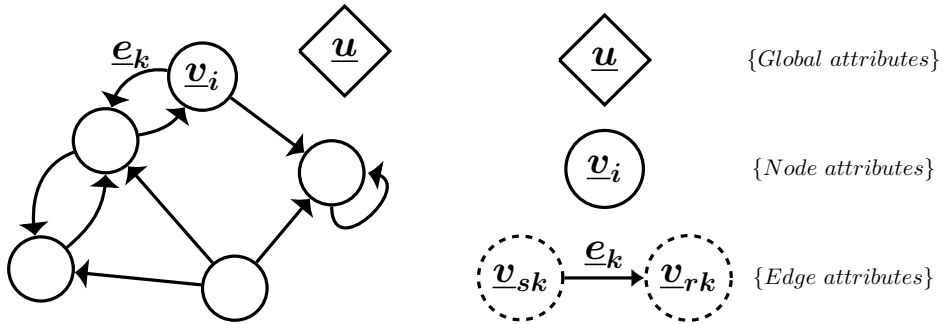

As discussed earlier, computations within the algorithm should be applicable to graphs of any size. In the most general case, the algorithm is required to take a graph
(with node, edge and global attributes represented by vectors) as input. A general graph of this type is shown in Figure \ref{fig:general_graph}; edges are directed from
one node to another; self-connections and  multi-graphs are allowed. For the sake of simplicity (and this is sufficient for the case study later), it is assumed that all
node attribute vectors have the same dimension, the same being true for edge attributes. In the most general case, every element of the graph has attributes; however, if
it is convenient for the user, a class of elements (e.g.\ edges) may not have any attributes.

The output of the algorithm is also a graph. Most commonly, it is a graph with the same topology but with different values of attributes on all its elements. However,
the attribute vectors of the output graph do not need to have the same dimension as those of the input graph. For example, the nodes in a graph representing a mass-spring
system moving in time might have input attributes specified by the forces applied on each node and the current displacements and velocities of the nodes, while the output
graph from the algorithm may only have attributes showing the velocities after the next time step of the simulation.

The main iterations of the graph network algorithm are divided into {\em computational blocks} or {\em GN blocks}; these are in turn divided into three updating operations
or steps, which define the transition from a given graph to another with different (or updated) values on the attribute vectors of its elements:

\begin{enumerate}
    \item the edge updates,
    \item the node updates, and
    \item the global attribute updates.
\end{enumerate}

In each iteration, or block, the values of the new attribute vectors are calculated for every element of the graph. The blocks are repeated until some convergence or
termination criterion is met. In the case of learning a model from data, the maximum number of computational blocks applied is a hyperparameter of the model, whose
definition may be informed by physical intuition into the problem, or established by cross validation. As described later, some of the individual updates within a block
may be omitted according to the application for which the model is used. A more detailed description of the individual updating steps follows. Quantities after updates are
denoted by the ' symbol.

\subsubsection{Edge update}
\label{sssec:eup}

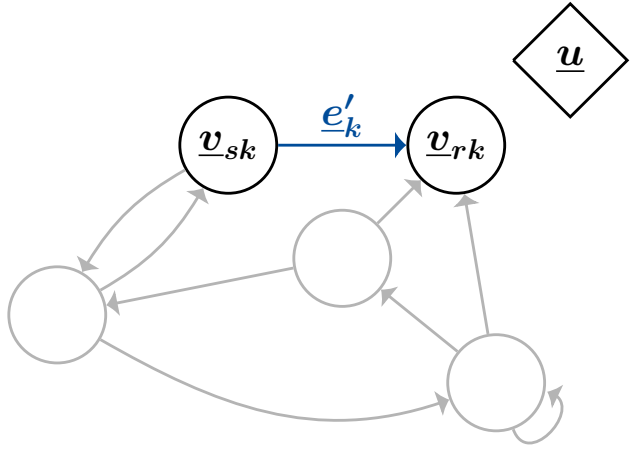
\begin{figure}[htbp!]
    \centering
    \scalebox{0.75}{
    \begin{tikzpicture}
        
\definecolor{gray1}{RGB}{180, 180, 180}
\definecolor{blue1}{RGB}{0, 76, 153}
\node[circle,draw, minimum size=1.7cm, line width=0.5mm, gray1] (A) at  (0.5, 0) {};
\node[circle,draw, minimum size=1.7cm, line width=0.5mm] (B) at  (3.5, 3) {\huge $\bm{{\underline{v}}_{sk}}$};
\node[circle,draw, minimum size=1.7cm, line width=0.5mm] (C) at  (7.5, 3) {\huge $\bm{{\underline{v}}_{rk}}$};
\node[circle,draw, minimum size=1.7cm, line width=0.5mm, gray1] (D) at  (5.5, 1.0) {};
\node[circle,draw, minimum size=1.7cm, line width=0.5mm, gray1] (E) at  (8.2, -1.2) {};

\draw[-{Latex[width=5mm, length=2mm]}, line width=0.5mm, gray1] (A) to[out=-30, in=200] (E);
\draw[-{Latex[width=5mm, length=2mm]}, line width=0.5mm, gray1] (A) to[out=30, in=240] (B);
\draw[-{Latex[width=5mm, length=2mm]}, line width=0.5mm, gray1] (B) to[out=210, in=60] (A);
\draw[-{Latex[width=5mm, length=2mm]}, line width=0.5mm, blue1] (B) to node[above] {\huge $\bm{{\underline{e}}'_{k}}$} (C);

\draw[line width=0.5mm,-{Latex[width=5mm, length=2mm]},shorten >=1pt, gray1] (E) to [out=290,in=-10,loop,looseness=2.4] (E);
\draw[-{Latex[width=5mm, length=2mm]}, line width=0.5mm, gray1] (D) to (A);
\draw[-{Latex[width=5mm, length=2mm]}, line width=0.5mm, gray1] (E) to (C);
\draw[-{Latex[width=5mm, length=2mm]}, line width=0.5mm, gray1] (E) to (D);
\draw[-{Latex[width=5mm, length=2mm]}, line width=0.5mm, gray1] (D) to (C);

\draw[line width=0.5mm] (8.5, 4.5) -- (9.5, 3.5) -- (10.5, 4.5) -- (9.5, 5.5) -- (8.5, 4.5);
\node[] at (9.5, 4.5) {\huge $\bm{\underline{u}}$};

\end{tikzpicture}
    }
    \caption{Edge update step.}
    \label{fig:first_update_step}
\end{figure}

During this step, illustrated in in Figure \ref{fig:first_update_step}, the attributes of the edges of the graph are updated. The new attributes of the edge (${\ue_k}'$) are computed using the initial attributes of: the edge itself, the sender and receiver nodes of the edge and the global attributes of the graph. The update takes the form,

\begin{equation}
     {\ue_k}' = \phi^e (\ue_k, \uv_{s_k}, \uv_{r_k}, \uu)
\label{eq:eup}
\end{equation}
where $\phi^e$ is a function {\em learned} during the overall training process.

Note that this operation is purely local, and is invariant under permutation operations on the graph; this follows because a change in the indexing of the nodes would
relabel the indices for the sender and receiver nodes. Furthermore, as the edge update is applied to {\em all} edges in this step, the overall step is permutation invariant.
The role of $\phi^e$ in this step is rather like the convolution kernel in a CNN, in the sense that the same function is applied locally across all edges.

\subsubsection{Node update}
\label{sssec:nup}

\begin{figure}[htbp!]
    \centering
    \scalebox{0.75}{
    \begin{tikzpicture}

\definecolor{gray1}{RGB}{180, 180, 180}
\definecolor{blue1}{RGB}{0, 76, 153}
\node[circle,draw, minimum size=1.7cm, line width=0.5mm, gray1] (A) at  (0.5, 0) {};
\node[circle,draw, minimum size=1.7cm, line width=0.5mm, gray1] (B) at  (3.5, 3) {};
\node[circle,draw, minimum size=1.7cm, line width=0.5mm, blue1] (C) at  (7.5, 3) {\huge $\bm{\underline{v}'_{k}}$};
\node[circle,draw, minimum size=1.7cm, line width=0.5mm, gray1] (D) at  (5.5, 1.0) {};
\node[circle,draw, minimum size=1.7cm, line width=0.5mm, gray1] (E) at  (8.2, -1.2) {};

\draw[-{Latex[width=5mm, length=2mm]}, line width=0.5mm, gray1] (A) to[out=-30, in=200] (E);
\draw[-{Latex[width=5mm, length=2mm]}, line width=0.5mm, gray1] (A) to[out=30, in=240] (B);
\draw[-{Latex[width=5mm, length=2mm]}, line width=0.5mm, gray1] (B) to[out=210, in=60] (A);
\draw[-{Latex[width=5mm, length=2mm]}, line width=0.5mm] (B) to node[above] {\huge $\bm{{\underline{e}}'_{k}}$} (C);

\draw[line width=0.5mm,-{Latex[width=5mm, length=2mm]},shorten >=1pt, gray1] (E) to [out=290,in=-10,loop,looseness=2.4] (E);
\draw[-{Latex[width=5mm, length=2mm]}, line width=0.5mm, gray1] (D) to (A);
\draw[-{Latex[width=5mm, length=2mm]}, line width=0.5mm] (E) to (C);
\draw[-{Latex[width=5mm, length=2mm]}, line width=0.5mm, gray1] (E) to (D);
\draw[-{Latex[width=5mm, length=2mm]}, line width=0.5mm] (D) to (C);

\draw[line width=0.5mm] (8.5, 4.5) -- (9.5, 3.5) -- (10.5, 4.5) -- (9.5, 5.5) -- (8.5, 4.5);
\node[] at (9.5, 4.5) {\huge $\bm{\underline{u}}$};

\end{tikzpicture}
    }
    \caption{Node update.}
    \label{fig:second_update_step}
\end{figure}
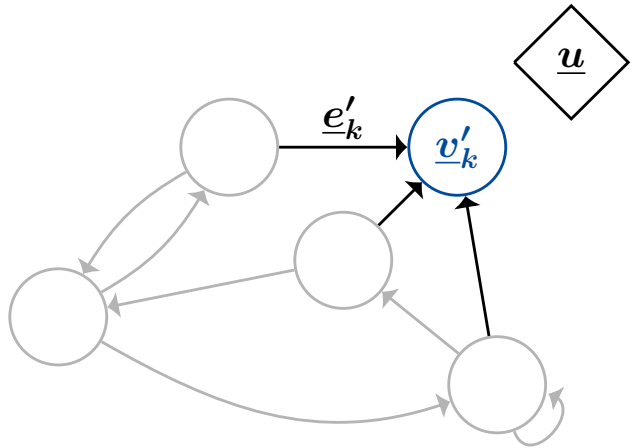

In the second step, illustrated in Figure \ref{fig:second_update_step}, the attributes of each node are updated; to do so, the attributes of all the edges $E'$, pointing
to the node to be updated, are fed into a aggregative function $\rho^{e \to v}$. This function must be a summation or averaging function, so that it takes as many inputs
as are dictated by the neighbourhood of the node and outputs an attribute vector of fixed size. In this step, $\rho^{e \to v}$ is analogous to the activation function in a
normal neural network and is not learned, but pre-specified. Following this step, another function $\phi^{v}$ is evaluated,

\begin{equation}
     {\uv_k}' = \phi^v ( \rho^{e \to v}(E'), \uv_k, \uu )
\label{eq:nup}
\end{equation}
with arguments from the aggregative function and the initial node and global attributes at the start of the block. It should be clear that this update is also permutation
invariant and local. The function $\phi^v$ is also learned from the training data.

\subsubsection{Global update}
\label{sssec:gup}

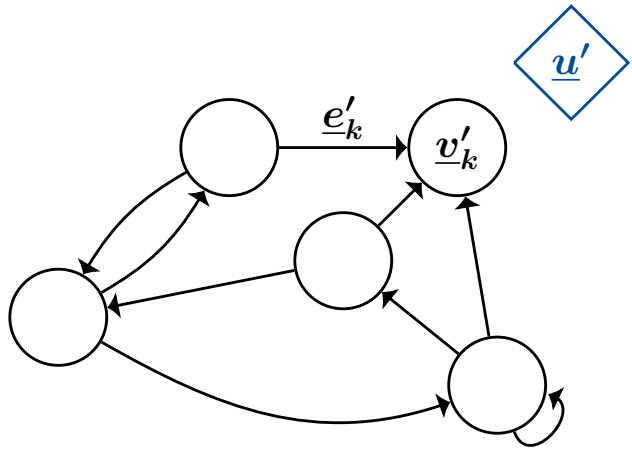
\begin{figure}[htbp!]
    \centering
    \scalebox{0.75}{
    \begin{tikzpicture}
        
\definecolor{gray1}{RGB}{180, 180, 180}
\definecolor{blue1}{RGB}{0, 76, 153}
\node[circle,draw, minimum size=1.7cm, line width=0.5mm] (A) at  (0.5, 0) {};
\node[circle,draw, minimum size=1.7cm, line width=0.5mm] (B) at  (3.5, 3) {};
\node[circle,draw, minimum size=1.7cm, line width=0.5mm] (C) at  (7.5, 3) {\huge $\bm{\underline{v}'_{k}}$};
\node[circle,draw, minimum size=1.7cm, line width=0.5mm] (D) at  (5.5, 1.0) {};
\node[circle,draw, minimum size=1.7cm, line width=0.5mm] (E) at  (8.2, -1.2) {};

\draw[-{Latex[width=5mm, length=2mm]}, line width=0.5mm] (A) to[out=-30, in=200] (E);
\draw[-{Latex[width=5mm, length=2mm]}, line width=0.5mm] (A) to[out=30, in=240] (B);
\draw[-{Latex[width=5mm, length=2mm]}, line width=0.5mm] (B) to[out=210, in=60] (A);
\draw[-{Latex[width=5mm, length=2mm]}, line width=0.5mm] (B) to node[above] {\huge $\bm{{\underline{e}}'_{k}}$} (C);

\draw[line width=0.5mm,-{Latex[width=5mm, length=2mm]},shorten >=1pt] (E) to [out=290,in=-10,loop,looseness=2.4] (E);
\draw[-{Latex[width=5mm, length=2mm]}, line width=0.5mm] (D) to (A);
\draw[-{Latex[width=5mm, length=2mm]}, line width=0.5mm] (E) to (C);
\draw[-{Latex[width=5mm, length=2mm]}, line width=0.5mm] (E) to (D);
\draw[-{Latex[width=5mm, length=2mm]}, line width=0.5mm] (D) to (C);

\draw[line width=0.5mm, blue1] (8.5, 4.5) -- (9.5, 3.5) -- (10.5, 4.5) -- (9.5, 5.5) -- (8.5, 4.5);
\node[blue1] at (9.5, 4.5) {\huge $\bm{\underline{u}'}$};

\end{tikzpicture}
    }
    \caption{Global update.}
    \label{fig:third_update_step}
\end{figure}

The final step, illustrated in Figure \ref{fig:third_update_step}, is performed to update the global features. This time, the attribute vector of every node is passed as
argument into another aggregative function $\rho^{v \to u}$, and similarly, every edge feature vector into an aggregative function $\rho^{e \to u}$. Finally, a further
function evaluation yields,

\begin{equation}
     {\uu}' = \phi^u ( \rho^{e \to u}(E'), \rho^{v \to u}(V'), \uu )
\label{eq:gup}
\end{equation}

The function $\phi^u$, is learned from the training data, while $\rho^{e \to u}$ and $\rho^{v \to u}$ are selected by the user.

It is clear that the individual updating steps pass on information only from the local neighbourhoods of the graph elements. For this reason, if the user wishes, more
computational blocks can be applied on the input graph before reaching the final graph, which has the target values as its features. Additional computational steps have
the effect of extending the region of influence of the updates beyond the immediate neighbourhoods of the elements concerned. Once the blocks have been applied, the other
functions $\phi^e$ etc.\ in the model are updated.

\subsection{Graph neural network training}

In general, the functions $\phi$ described above could be represented by any nonparametric model; in practice, they are often selected to be neural networks.
Adding in the weights of these networks to the complete set of trainable parameters, completes the specification of the GN, which can now be referred to as a {\em Graph
Neural Network} (GNN). Training of the network parameters is now accomplished using backpropagation of errors via any classical ANN algorithm e.g.\ scaled conjugate
gradients \cite{Bishop,Bishop2}.

\begin{figure}[htbp!]
    \centering
    \scalebox{0.85}{
    \begin{tikzpicture}
        
\definecolor{gray2}{RGB}{160, 160, 160}
\draw[line width=0.5mm] (0, 0) to (0, 8);
\draw[line width=0.5mm] (0, 8) to (10, 8);
\draw[line width=0.5mm] (10, 8) to (10, 0);
\draw[line width=0.5mm] (10, 0) to (0, 0);

\node[] at (-0.7, 7.2) {\huge $\bm{\underline{u}}$};
\node[] at (-0.7, 4.0) {\huge $\bm{\underline{v}}$};
\node[] at (-0.7, 0.8) {\huge $\bm{\underline{e}}$};

\node[] at (10.7, 7.2) {\huge $\bm{\underline{u}'}$};
\node[] at (10.7, 4.0) {\huge $\bm{\underline{v}'}$};
\node[] at (10.7, 0.8) {\huge $\bm{\underline{e}'}$};

\node[circle,draw, minimum size=0.6cm, line width=0.5mm] (A) at  (8.4, 7.2) {$\bm{\phi^u}$};

\draw[-{Latex[width=5mm, length=2mm]}, line width=0.5mm] (-0.2, 7.2) to (A);
\draw[-{Latex[width=5mm, length=2mm]}, line width=0.5mm] (A) to (10.3, 7.2);

\node[circle,draw, minimum size=1.0cm, line width=0.5mm] (B) at  (5.0, 4.0) {$\bm{\phi^u}$};
\draw[-{Latex[width=5mm, length=2mm]}, line width=0.5mm] (-0.2, 7.2) to (B);
\draw[-{Latex[width=5mm, length=2mm]}, line width=0.5mm] (-0.2, 4.0) to (B);
\draw[-{Latex[width=5mm, length=2mm]}, line width=0.5mm] (B) to (10.3, 4.0);

\node (rho3) at (7.2, 5.7) [draw,line width=0.5mm] {$\bm{\rho^{v \to u}}$};
\draw[-{Latex[width=5mm, length=2mm]}, line width=0.5mm] (B) to[out=30, in=270] (rho3);
\draw[-{Latex[width=5mm, length=2mm]}, line width=0.5mm] (rho3) to[out=60, in=210] (A);

\node[circle,draw, minimum size=1.0cm, line width=0.5mm] (C) at  (1.3, 0.8) {$\bm{\phi^e}$};
\draw[-{Latex[width=5mm, length=2mm]}, line width=0.5mm] (-0.2, 0.8) to (C);
\draw[-{Latex[width=5mm, length=2mm]}, line width=0.5mm] (C) to (10.3, 0.8);

\node (rho1) at (3.3, 2.0) [draw,line width=0.5mm] {$\bm{\rho^{e \to v}}$};
\draw[-{Latex[width=5mm, length=2mm]}, line width=0.5mm] (C) to[out=30, in=270] (rho1);
\draw[-{Latex[width=5mm, length=2mm]}, line width=0.5mm] (rho1) to[out=60, in=210] (B);

\node (rho2) at (8.0, 2.0) [draw,line width=0.5mm] {$\bm{\rho^{e \to u}}$};
\draw[-{Latex[width=5mm, length=2mm]}, line width=0.5mm] (C) to[out=5, in=210] (rho2);
\draw[-{Latex[width=5mm, length=2mm]}, line width=0.5mm] (rho2) to[out=70, in=270] (A);

\draw[line width=0.5mm] (2.0, 8.0) to (2.0, 0.0);
\draw[line width=0.5mm, gray2] (2.0, 0.0) to (2.0, -1.0);
\node[] at (1.0, -0.5) {\shortstack{Edge\\block}};

\draw[line width=0.5mm] (6.0, 8.0) to (6.0, 0.0);
\draw[line width=0.5mm, gray2] (6.0, 0.0) to (6.0, -1.0);
\node[] at (4.0, -0.5) {\shortstack{Node\\block}};

\node[] at (8.0, -0.5) {\shortstack{Global\\block}};

\end{tikzpicture}
    }
    \caption{Full computation block (motivated by \cite{Battaglia}).}
    \label{fig:full_gnn_block}
\end{figure}
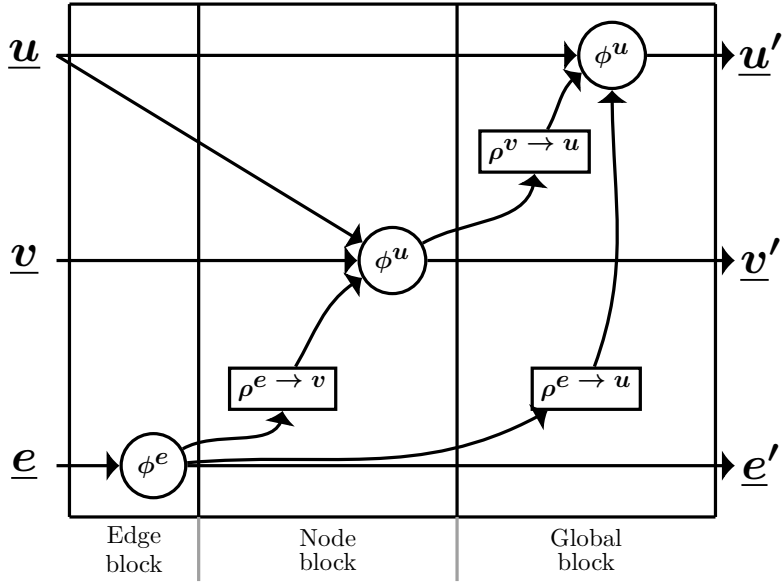

The full training iteration including the computational block is illustrated in Figure \ref{fig:full_gnn_block}. The gradients from an error function calculated according
to target values ($u'_{target}, V'_{target}, e'_{target}$) and the outputs ($u', V', e'$) are computed and backpropagated.

\begin{figure}[htbp!]
    \centering
    \scalebox{0.85}{
    \begin{tikzpicture}
        
\definecolor{gray2}{RGB}{160, 160, 160}
\draw[line width=0.5mm] (0, 0) to (0, 8);
\draw[line width=0.5mm] (0, 8) to (10, 8);
\draw[line width=0.5mm] (10, 8) to (10, 0);
\draw[line width=0.5mm] (10, 0) to (0, 0);

\node[] at (-0.7, 7.2) {\huge $\bm{\underline{u}}$};
\node[] at (-0.7, 4.0) {\huge $\bm{\underline{v}}$};
\node[] at (-0.7, 0.8) {\huge $\bm{\underline{e}}$};

\node[] at (10.7, 4.0) {\huge $\bm{\underline{v}'}$};

\node[circle,draw, minimum size=1.0cm, line width=0.5mm] (B) at  (5.0, 4.0) {$\bm{\phi^u}$};
\draw[-{Latex[width=5mm, length=2mm]}, line width=0.5mm] (-0.2, 7.2) to (B);
\draw[-{Latex[width=5mm, length=2mm]}, line width=0.5mm] (-0.2, 4.0) to (B);
\draw[-{Latex[width=5mm, length=2mm]}, line width=0.5mm] (B) to (10.3, 4.0);

\node[circle,draw, minimum size=1.0cm, line width=0.5mm] (C) at  (1.3, 0.8) {$\bm{\phi^e}$};
\draw[-{Latex[width=5mm, length=2mm]}, line width=0.5mm] (-0.2, 0.8) to (C);

\node (rho1) at (3.3, 2.0) [draw,line width=0.5mm] {$\bm{\rho^{e \to v}}$};
\draw[-{Latex[width=5mm, length=2mm]}, line width=0.5mm] (C) to[out=30, in=270] (rho1);
\draw[-{Latex[width=5mm, length=2mm]}, line width=0.5mm] (rho1) to[out=60, in=210] (B);

\draw[line width=0.5mm] (2.0, 8.0) to (2.0, 0.0);
\draw[line width=0.5mm, gray2] (2.0, 0.0) to (2.0, -1.0);
\node[] at (1.0, -0.5) {\shortstack{Edge\\block}};

\draw[line width=0.5mm] (6.0, 8.0) to (6.0, 0.0);
\draw[line width=0.5mm, gray2] (6.0, 0.0) to (6.0, -1.0);
\node[] at (4.0, -0.5) {\shortstack{Node\\block}};

\node[] at (8.0, -0.5) {\shortstack{Global\\block}};

\end{tikzpicture}
    }
    \caption{Reduced computation block.}
    \label{fig:reduced_gnn_block}
\end{figure}
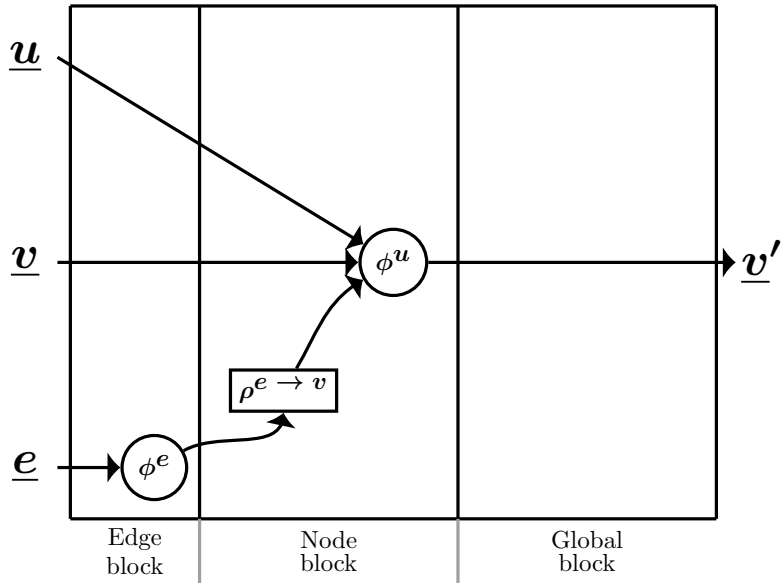

With this approach to training the model, several target values may be approximated. The output graph may have values for the attributes of nodes and edges and the global
ones. In practice, the user may only be interested in approximating attributes on one type of element, or even just a subset of those elements. If, for example, only the
node attributes are of predictive interest (e.g.\ the material and geometric properties of irreducible elements in PBSHM \cite{PBSHMMSSP2}), a much reduced computational
block is possible, as depicted in Figure \ref{fig:reduced_gnn_block}.

\section{A GNN Application in Population-Based SHM}
\label{sec:gnn_application}

\subsection{Application description}

\begin{figure}[htbp!]
    \centering
    \includegraphics[width=.90\textwidth]{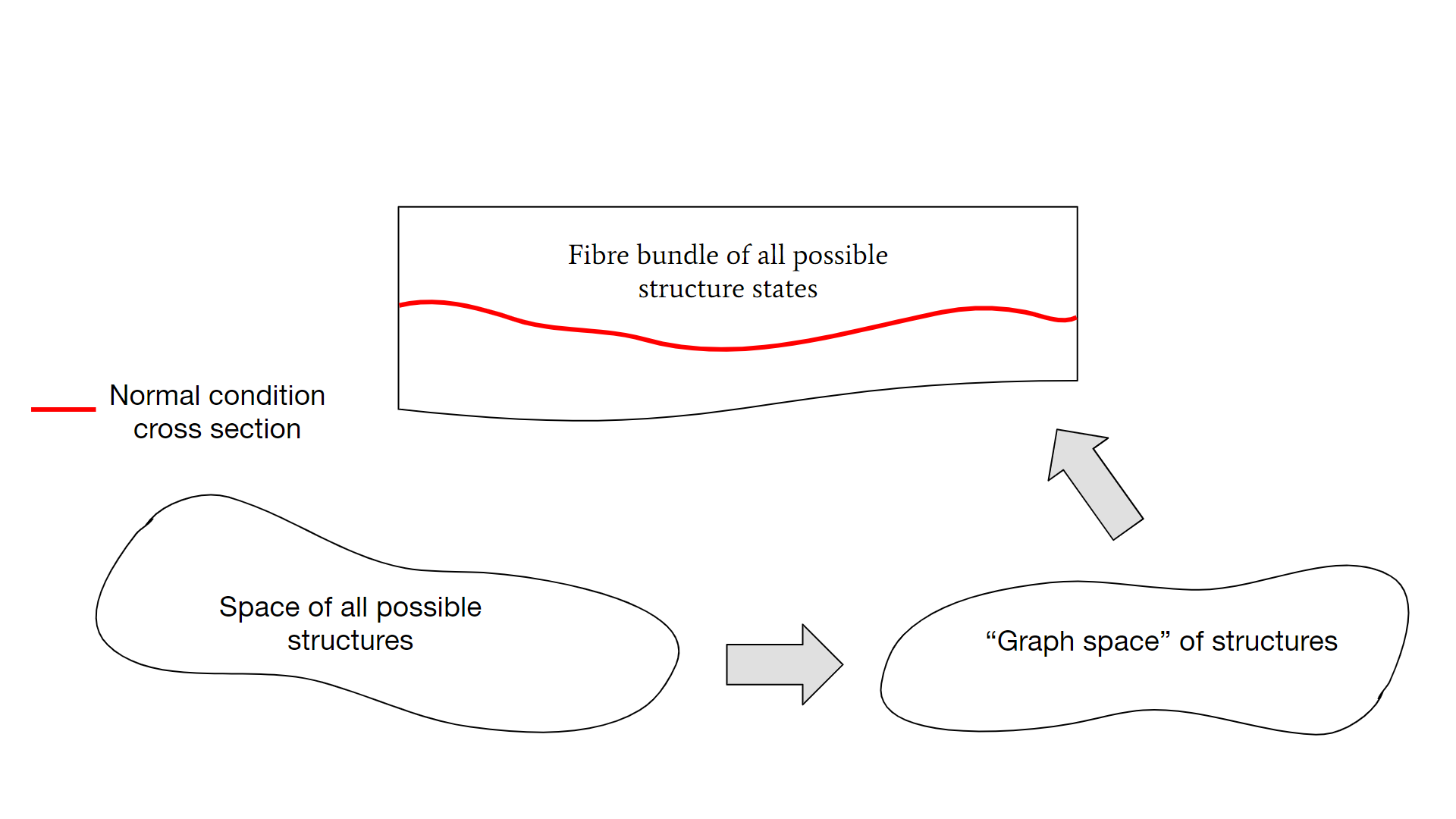}
    \caption{Scheme to be followed to approximate first natural frequencies of structures using GNNs.}
    \label{fig:algorithm_scheme}
\end{figure}

The application problem chosen here is the estimation of the normal cross section of the feature bundle over a population of structures expressed as attributed graphs. To
simplify matters, the feature will be the first natural frequency of the structures in question; this is still a highly nontrivial function of the structural composition.
Because the structures exist within an abstract (non-vector) space which is not immediately expressed as a manifold, the GNN algorithm is employed, since it is a non-Euclidean algorithm. The procedure implicitly followed is shown in Figure \ref{fig:algorithm_scheme}. As discussed earlier, there is no need of a direct embedding of the space of graphs. Before the GNN algorithm is applied, there is a need to convert the structures of interest into attributed graphs (AGs); a general approach to this problem is explained in the
second paper in this series \cite{PBSHMMSSP2}. This step will also be simplified in the current work by concentrating on a population of structures -- {\em trusses} -- which
have a direct and unambiguous conversion into AGs.

\begin{figure}[htbp!]
    \centering
    \includegraphics[width=0.75\textwidth]{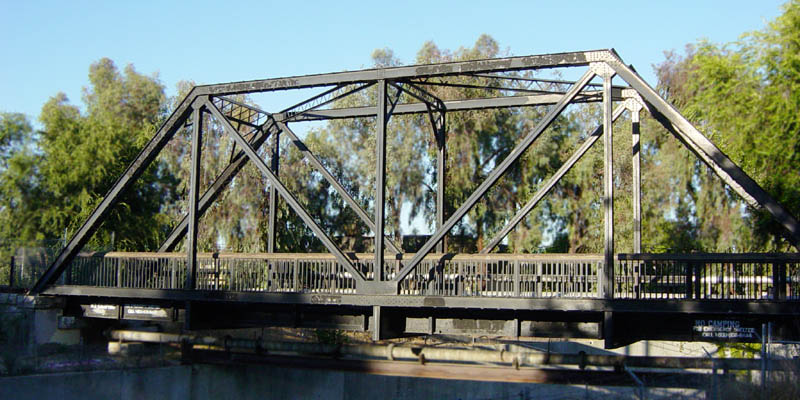}
    \\ (a) \\ \vspace*{3mm}
    \includegraphics[width=0.75\textwidth]{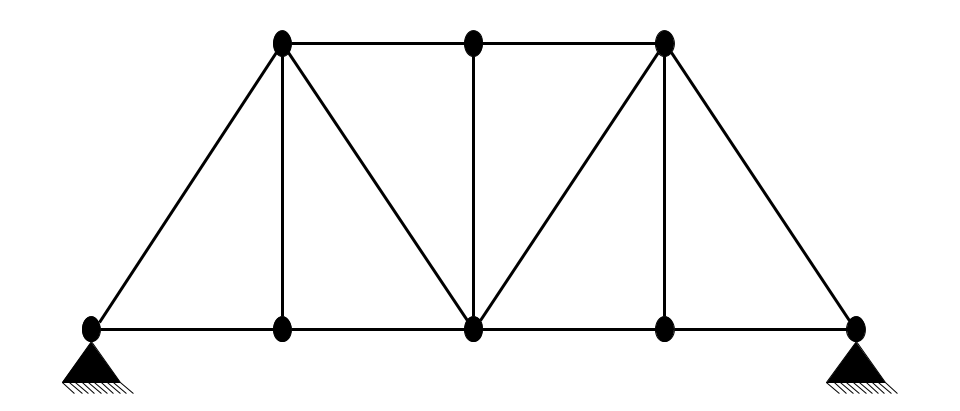}
    \\ (b)
    \caption{(a) Truss railway bridge, and (b) simple planar model.}
    \label{fig:railway_bridge}
\end{figure}

Truss structures are assembled from rod elements connected at simple joints; the members are connected only at their edges and so are `two force elements', loaded only in
their axial dimension. As a result, the elements are either under simple tension or compression. Despite their apparent simplicity, truss structures are widely used in
engineering, with both planar and three-dimensional geometries. Many bridges are constructed as trusses, particularly for railway use, as well as stadium rooftops, antennae, cranes etc. In Figure \ref{fig:railway_bridge}, a railway bridge and a simple model representation are shown\footnote{The bridge image was taken from
\url{https://en.wikipedia.org/wiki/Truss}: last accessed 11/11/20.}. Considering the planar model, it is clear that the transformation of trusses into graphs should be a
straightforward task.

In forming the AG models of trusses, it is useful to depart from the general principles for forming irreducible element (IE) models as discussed in \cite{PBSHMMSSP2}, where
one would identify the truss members as $[\mbox{rod}]$ IEs\footnote{Square brackets are used to indicate irreducible element classes \cite{PBSHMMSSP2}.} and regard the
nodes of the truss as the joints. Because the truss has such a natural representation as a graph,
it is convenient to identify the nodes as the IEs and the edges as the joints. The attributes for the nodes will then be the positions and boundary conditions of the
vertices, while the attributes of the edges will be the material and physical characteristics of the rod members. Potential environmental conditions, such as temperature
can also be encoded within the graph as global attributes. The attribute vectors will have two real values for the two-dimensional coordinates and two values with
binary encoding defining whether the node is fixed in the $x$ or $y$ directions respectively. Internal connections between members are modelled as pinned. The physical
attributes of the edges/members that will affect the problem of defining natural frequencies, are the stiffnesses of the members. The stiffness $K$ of a truss member is
given by $K = EA/L$, where $E$ is the Young's modulus of the member, $A$ is its area and $L$ its length. The length and the sine and cosine of the members' angles could
actually be inferred by the algorithm from the coordinates of the nodes, but they are included here in the edge attributes to facilitate the algorithm in defining the
member stiffnesses.

\begin{figure}[htbp!]
\centering
    \begin{subfigure}[b]{\textwidth}
    \centering
    \scalebox{0.9}{
    \begin{tikzpicture}

\node[circle,draw=black, fill=white, label=below:\textcolor{blue}{$[x, y, 1, 1]$}] (a) at (0, 0) {};

\node[circle,draw=black, fill=white, label=below:\textcolor{blue}{$[x, y, 0, 0]$}] (b) at (3, 0) {};

\node[circle,draw=black, fill=white, label=below:\textcolor{blue}{$[x, y, 0, 0]$}] (c) at (6, 0) {};

\node[circle,draw=black, fill=white, label=below:\textcolor{blue}{$[x, y, 0, 0]$}] (d) at (9, 0) {};

\node[circle,draw=black, fill=white, label=below:\textcolor{blue}{$[x, y, 1, 1]$}] (e) at (12, 0) {};

\node[circle,draw=black, fill=white, label=above:\textcolor{blue}{$[x, y, 0, 0]$}] (f) at (3, 4) {};

\node[circle,draw=black, fill=white, label=above:\textcolor{blue}{$[x, y, 0, 0]$}] (g) at (6, 4) {};

\node[circle,draw=black, fill=white, label=above:\textcolor{blue}{$[x, y, 0, 0]$}] (h) at (9, 4) {};

\draw[thick] (a) -- (b) node[midway, below, text=orange] {$[0, 0, 1]$};
\draw[thick] (b) -- (c) node[midway, below, text=orange] {$[0, 0, 1]$};
\draw[thick] (c) -- (d) node[midway, below, text=orange] {$[0, 0, 1]$};
\draw[thick] (d) -- (e) node[midway, below, text=orange] {$[0, 0, 1]$};
\draw[thick] (a) -- (f) node[midway, above, sloped, text=orange] {$[0, 1, 0]$};
\draw[thick] (f) -- (g) node[midway, above, sloped, text=orange] {$[1, 0, 0]$};
\draw[thick] (g) -- (h) node[midway, above, sloped, text=orange] {$[1, 0, 0]$};
\draw[thick] (b) -- (f) node[midway, above, sloped, text=orange] {$[0, 1, 0]$};
\draw[thick] (f) -- (c) node[midway, above, sloped, text=orange] {$[0, 1, 0]$};
\draw[thick] (c) -- (g) node[midway, above, sloped, text=orange] {$[0, 1, 0]$};
\draw[thick] (c) -- (h) node[midway, above, sloped, text=orange] {$[0, 1, 0]$};
\draw[thick] (e) -- (h) node[midway, above, sloped, text=orange] {$[0, 1, 0]$};
\draw[thick] (d) -- (h) node[midway, above, sloped, text=orange] {$[0, 1, 0]$};

\node[text=black!60!green] at (1.5, 5) {$[30]$};

\end{tikzpicture}
    }
    \caption{Detailed attribute representation.}
    \end{subfigure}
    \begin{subfigure}[b]{\textwidth}
    \centering
    \scalebox{0.9}{
    \begin{tikzpicture}

\definecolor{purple1}{RGB}{102, 0, 204}
\definecolor{red1}{RGB}{204, 0, 0}
\definecolor{orange1}{RGB}{255, 128, 0}

\node[circle,draw=black, fill=black] (a) at (0, 0) {};

\node[circle,draw=black, fill=white] (b) at (3, 0) {};

\node[circle,draw=black, fill=white] (c) at (6, 0) {};

\node[circle,draw=black, fill=white] (d) at (9, 0) {};

\node[circle,draw=black, fill=black] (e) at (12, 0) {};

\node[circle,draw=black, fill=white] (f) at (3, 4) {};

\node[circle,draw=black, fill=white] (g) at (6, 4) {};

\node[circle,draw=black, fill=white] (h) at (9, 4) {};

\draw[line width=0.6mm, purple1] (a) -- (b);
\draw[line width=0.6mm, purple1] (b) -- (c);
\draw[line width=0.6mm, purple1] (c) -- (d);
\draw[line width=0.6mm, purple1] (d) -- (e);
\draw[line width=0.6mm, red1] (a) -- (f);
\draw[line width=0.6mm, red1] (b) -- (f);
\draw[line width=0.6mm, red1] (f) -- (c);
\draw[line width=0.6mm, red1] (c) -- (g);
\draw[line width=0.6mm, red1] (c) -- (h);
\draw[line width=0.6mm, red1] (e) -- (h);
\draw[line width=0.6mm, red1] (d) -- (h);
\draw[line width=0.6mm, orange1] (f) -- (g);
\draw[line width=0.6mm, orange1] (g) -- (h);

\node[text=black!60!green] at (1.5, 5) {$[30]$};

\end{tikzpicture}
    }
    \caption{Simplified `categorical' representation.}
    \end{subfigure}
\caption{Network representation of the bridge shown in Figure \ref{fig:railway_bridge}. (a) shows a detailed representation in terms or explicit node (blue) and edge (red)
         attributes; (b) shows a simplified representation indicating fixed (black) and pinned (white) nodes, with the three member types colour coded as `top' (orange),
         `middle' (red) and `bottom' (blue).}
\label{fig:bridge_network_representation}
\end{figure}
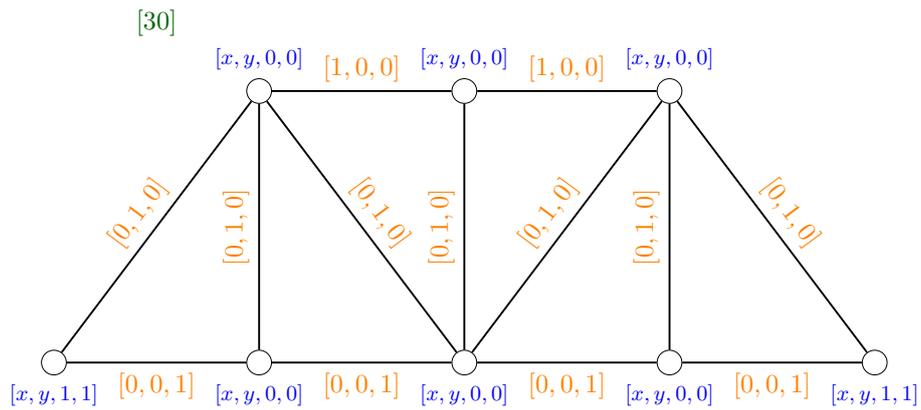
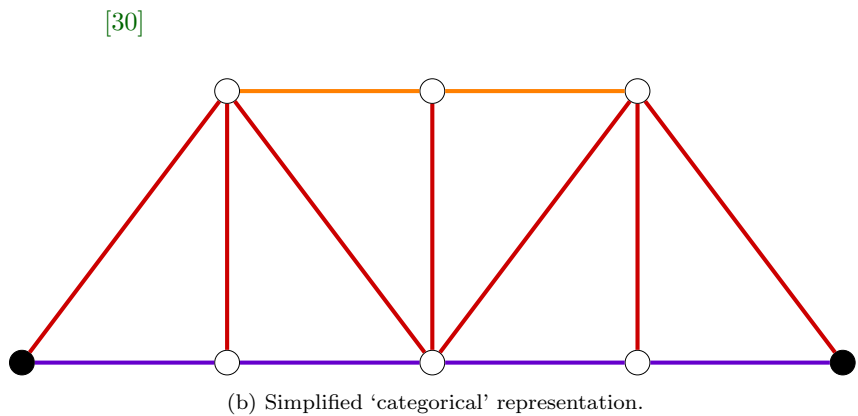

Another way to include the information needed to infer member stiffnesses is by a {\em categorical encoding}. Truss members could be categorised according to their
material and cross-sectional area; different members have the same material and cross section and so will belong to the same category. Different categories may be defined
and the category to which an edge/member belongs can be specified via several binary variables. For example, considering the truss structure and model from Figure \ref{fig:railway_bridge}, one might assume that the members at the top of the structure have the same material and cross section, that the members connecting the top and
bottom of the structure belong to a second class and that the members at the bottom to a third one. Under these assumptions, the AG representation of the structure is shown
in Figure \ref{fig:bridge_network_representation}. The figure shows that the two nodes on the left and right side at the base of the structure have fixed boundary
conditions, while other nodes are free to move (components 3 and 4 of the node attribute vectors). The binary attribute vectors associated with edges specify the members
as belonging to the {\em top}, {\em centre} or {\em bottom} classes. A potential global attribute of the graph, is the environmental temperature ($30^{\circ}$C) that the
truss is experiencing; this is shown in the top left of the figure.

\subsection{Simulation Experiments}
\label{subsec:simex}

Using the graphical representation of trusses discussed above, three studies were carried out in order to investigate the capability of the GNN algorithm in estimating the
normal section across a population. As mentioned earlier, the feature of interest is the first natural frequency (denoted $\omega$ here) of the structures. Each case study considers a different population of trusses.

\begin{figure}[htbp!]
    \centering
    \includegraphics[width=.77\textwidth]{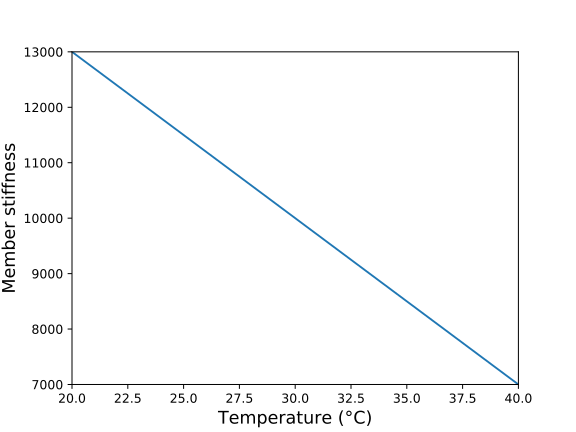}
    \caption{Linear relationship between temperature and $EA$ of first type of members (Case Studies Two and Three). Note that it is clearly unrealistic to assume
    that the stiffness of a member would halve over a range of 20 degrees Celcius. However, the point of the exercise here is to demonstrate that the methodology
    works well, even in the presence of large confounding influences. For this reason, the influence of temperature is deliberately exaggerated. We thank one of
    the anonymous reviewers for pointing out that clarification was necessary here.}
    \label{fig:member_1_EA}
\end{figure}

\begin{figure}[htbp!]
    \centering
    \includegraphics[width=.77\textwidth]{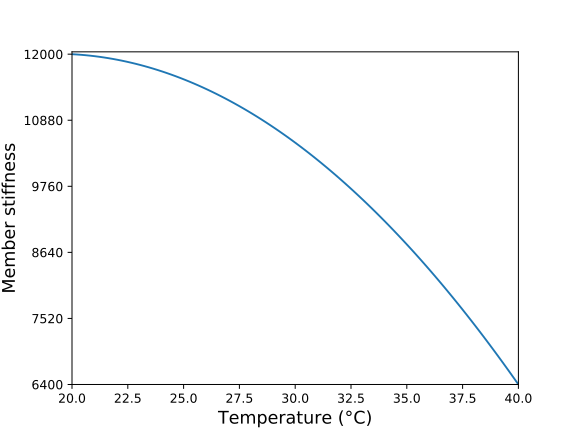}
    \caption{Nonlinear relationship between temperature and $EA$ of second type of members (Case Study Three).}
    \label{fig:member_2_EA}
\end{figure}

\begin{enumerate}
\item {\bf Case Study One}: The first study concerns a truss population represented by graphs with one common type of member, with defined area $A$ and Young's modulus $E$.
      In combination, the quantity $EA$ is equal to $10^4$ and there is no temperature variation in this study. Because the structures can have different numbers of
      elements and differing connectivity, they represent a truly heterogeneous population \cite{PBSHMMSSP2,PBSHMMSSP3}; in fact, it is highly heterogeneous in terms of
      topology. As there is only type of member in this example, there is no need for the categorical encoding in the edge attributes. \\

\item {\bf Case Study Two}: In this study the members have a linear relationship between the temperature and their $EA$, which is shown in Figure \ref{fig:member_1_EA}. In
      this case, the quantity $EA$ is continuously variable according to the same function for every member; therefore, as in the first case study, the edge attributes do
      not need a categorical/binary encoding of the member type. \\

\item {\bf Case Study Three}: In this case, the population allows a second type of member which has a nonlinear relationship between $EA$ and global temperature; thus
      requiring a binary encoding of the member type. This relationship is shown in Figure \ref{fig:member_2_EA} ($EA = -13T^{2} + 500T + 7200$, where $T$ is the temperature).
\end{enumerate}

In all the studies here, the populations of two-dimensional trusses were created by selecting the number of nodes randomly in the interval $[10, 40]$, with coordinates
$x$ and $y$ in the interval $[0, 10]$ and then creating the graphs by applying Delaunay triangulation \cite{Delaunay}, on the nodes. Nodes were fixed in the $x$ and $y$
directions randomly for all three populations. Random types of member were assigned to every edge of the graph regarding the third case study. Random temperatures in the
interval $[20, 40]$ were assigned to each truss for the second and third case studies. In all cases, the first natural frequency was calculated from an eigenvalue analysis
of the truss stiffness and mass matrices. In order to facilitate training by providing more information to the algorithm, the {\em sine} and {\em cosine} of angles and length of
each member -- encoding the relative positions of the nodes -- were also included in the edge attributes.

\subsection{Results: Case Study One}
\label{subsec:res1}

\begin{figure}[htbp!]
    \centering
    \includegraphics[width=0.35\textwidth]{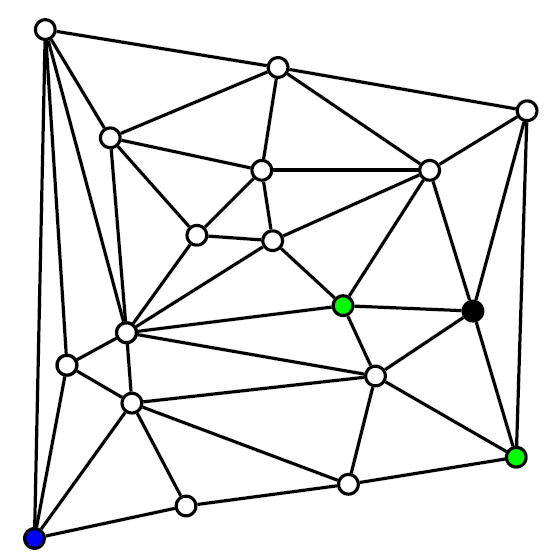}
    \\ (a) \\ \vspace*{3mm}
    \includegraphics[width=0.35\textwidth]{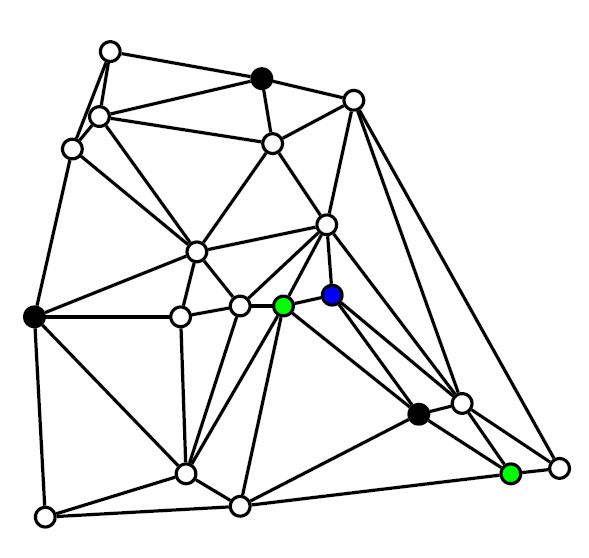}
    \\ (b) \\ \vspace*{3mm}
    \includegraphics[width=0.35\textwidth]{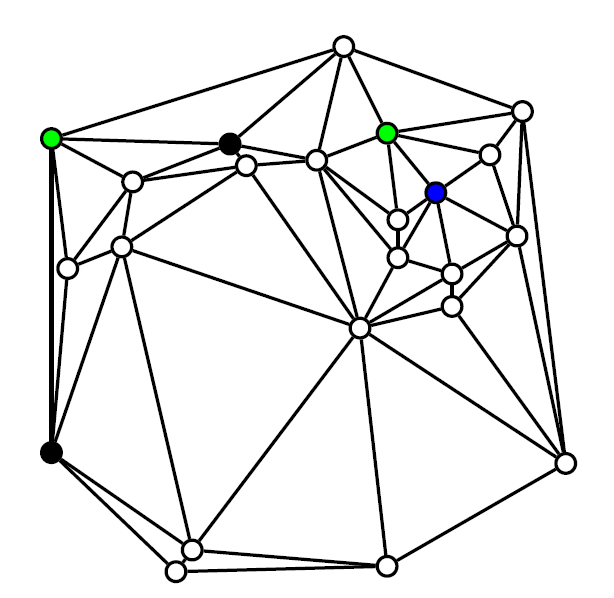}
    \\ (c)
    \caption{Three randomly-selected truss structures generated as part of the training data for the GNN algorithm in Case Study One. Black nodes represent
             fully fixed nodes, green nodes are fixed in the $x$ direction, blue are fixed in the $y$ direction and in white are free (pinned) nodes).}
    \label{fig:randtr}
\end{figure}

For each experiment, three datasets (training, validation and testing) were created, comprising data for 16000 trusses each; Figure \ref{fig:randtr} shows three
randomly-selected structures from the data set. During training, different GNN model structures were considered and the a cross-validation approach was used to select the
optimum. This exercise established that three computation blocks gave the best validation accuracy. Cross-validation was also used to determine the size of the neural
networks used in the edge, node and global updates.

Cross validation presented a computational problem for the current study. In order to establish network hyperparameters like the number of units per hidden layer, it is
standard practice to adopt an approach like Tarassenko's \cite{Tarassenko}, whereby the model is trained many times over a large range of hyperparameters and evaluated on
the validation data; the `optimal' hyperparameters are selected as those which produce the smallest validation error. Unfortunately, many modern learning algorithms are
extremely computationally expensive;
in this case, single runs of the GNN algorithm took
several hours on a powerful desktop computer. As a result, it was infeasible to conduct an exhaustive search over network hyperparameters, and a more restricted search over
a number of possibilities was adopted in each case. It is important to note that this does not mean that the principle of cross validation was at all violated; the structures
were selected on the basis of their validation error. The only negative consequence of adopting a limited candidate set for the hyperparameters is that the study will not necessarily attain the best possible performance for models; however, it will be seen that the performance achieved was very acceptable.

A standard least-squares loss function was used to compute the validation error; however, in order to judge the results, a normalised mean-square error is reported, as
defined by,

\begin{equation}
    NMSE = \frac{100}{N \sigma_{\omega}^{2}}\sum_{i=1}^{N}(\hat{\omega_i} - \omega_i)^{2}
    \label{eq:NMSE}
\end{equation}
where $\omega_i$ is the target value for the natural frequency and $\hat{\omega}_i$ is the value estimated by the algorithm, $\sigma^2_{\omega}$ is the variance of $\omega$
over the data set concerned. This $NMSE$ is useful as a metric since it is equal to $100\%$ if the model predictions ($\hat{\omega}_i$) are set to the mean value, i.e.\
$\hat{\omega}_i = \overline{\omega}$; values lower than $100\%$ reveal that the model is indeed capturing correlations in the data. Experience with this NMSE indicates that
good models are obtained for values of less than 5\%, with a value of less than 1\% for excellent models.

\subsubsection{Mean aggregative function}
\label{sssec:maf}

Using the approach proposed in \cite{Battaglia}, a summation or an averaging function is specified in the GNN framework for the functions $\rho^{e \to v}$ etc. For the first
case study, this approach was used. Different numbers of computational blocks (CBs) and sizes of neural networks were tried incrementally. A fairly coarse search was applied
with about 10 random initialisations performed per architecture. An interesting observation was that as the sizes of neural networks in the initial CB were increased, the sizes
of the neural networks in later CBs also needed to be increased in order to achieve acceptable accuracy. Based on this observation, a gradual increase was implemented in the neural networks of the first CB, followed by proportional increases in the layer sizes of all the other networks.

The sizes of the networks in the first CB were tested using 20 to 600 units with an increment of 20. Following the scheme described, the best model was found to have three
computational blocks with neural network (NN) sizes as shown in Table \ref{Tab:GNN_arch_1}. The numbers represent the size of all the layers in the neural networks. For
example, in the first CB, the array $3, 64, 32$ represents a neural network with an three-node input layer, a 64-node hidden layer and a 32-node output layer. The input
size of some networks is fully defined by the size of the output layers of networks feeding them data, as shown schematically by Figure \ref{fig:full_gnn_block}. Such an
example is the edge neural network of the third CB which has an input layer with 290 nodes; this is the result of the concatenation of the updated edge attributes (50)
coming from the edge NN of the first CB, the updated node attributes ($2\times70$) (sender and receiver nodes) and the updated global attributes of the second CB (100).
The input layers of the first CB have size equal to the total number of node, edge and global attributes respectively, according to the encoding of the trusses described. Therefore, the input layer of the edge NN in the first CB is of size three (length, cosine of the member angle, sine of the member angle). All neural networks adopted use a {\em rectified linear} or {\em ReLU} activation function \cite{Geron}.

\begin{figure}[htbp!]
    \centering
    \includegraphics[width=.85\textwidth]{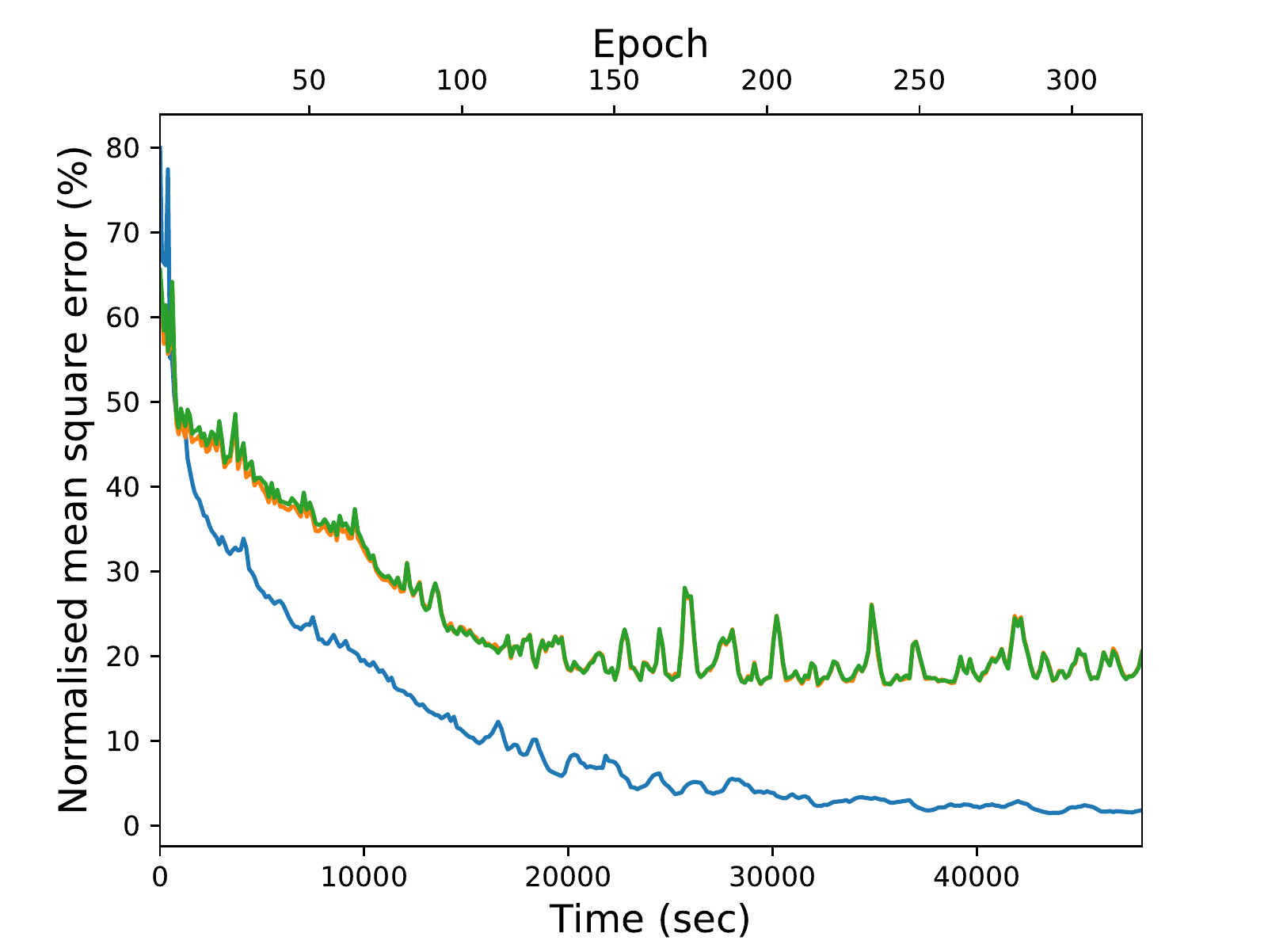}
    \caption{Training history of model using the summation aggregative function, training (blue), validation (orange) and testing (green) datasets: first case study.}
    \label{fig:sum_aggr_training_history}
\end{figure}

The training history in terms of {\em normalised mean square error} of the model is shown in Figure \ref{fig:sum_aggr_training_history}.

\begin{table}[htbp!]
\centering
\begin{tabular}{|l||l|l|l|}
\hline
		      & First CB & Second CB & Third CB \\
\hline
	Edge NN   &  3, 64, 32 &  132, 80,  50 & 290, 180, 80 \\
\hline
	Node NN   &  4, 72, 50 & 100, 100,  70 & 250, 300, 72 \\
\hline
    Global NN &  - & 120, 200, 100 & 252, 300, 72, 1 \\
\hline
\end{tabular}
\caption{Sizes of neural networks used in the GNN model with mean aggregative function: first case study.}
\label{Tab:GNN_arch_1}
\end{table}

The results gave a very low NMSE of 1.55\% on the training data; however the errors on the validation and testing sets were 17.16\% and 17.20\% respectively -- quite a bit
higher. It is encouraging that the validation and test errors are equal; this indicates a consistent generalisation and may indicate that the underlying physics is being
learnt by the algorithm. The validation and test errors are consistent with each other throughout training. This consistency shows that the cross-validation
strategy is working as intended; the degree of generalisation seen over the validation set is the same as that observed over the testing set. The algorithm achieves the
minimum validation error at around 140 epochs and 22000 seconds (6.1 hours). One epoch is considered to be the computation of all three CBs and update of all neural network parameters for every sample of the dataset.

One concern with the algorithm was that there may a lot of information lost because of the choice of aggregative function (averaging or summing). This possibility was
investigated further.

\subsubsection{Augmented aggregative function}
\label{sssec:aaf}

The next run of the algorithm tried to allow more information to flow through the neural networks via the aggregative functions in order to achieve lower error in predicting
the natural frequency. An augmentation of the aggregative functions was considered. If the aggregative functions are simply averaging or summing vectors, then information
about the distribution of these vectors is lost. The augmentation should provide more information about this distribution, so computing the variance as well as the mean value
was considered. Both the mean values and variances of the vectors are passed on to the neural networks of the GNN.

\begin{table}[htbp!]
\centering
\begin{tabular}{|l||l|l|l|}
\hline
	 ~   & First CB & Second CB & Third CB\\
\hline
    Edge NN     & 3, 64, 32 & 132, 80, 50   & 290, 180, 80 \\
\hline
    Node NN   & 4, 72, 50  & 150, 100, 70  & 330, 300, 72 \\
\hline
	Global NN   & - & 240, 200, 100  & 404, 450, 150, 1 \\
\hline
\end{tabular}
\caption{Sizes of neural networks used in the GNN model with the augmented aggregative function first case study.}
\label{Tab:GNN_arch_2}
\end{table}

\begin{figure}[htbp!]
    \centering
    \includegraphics[width=.85\textwidth]{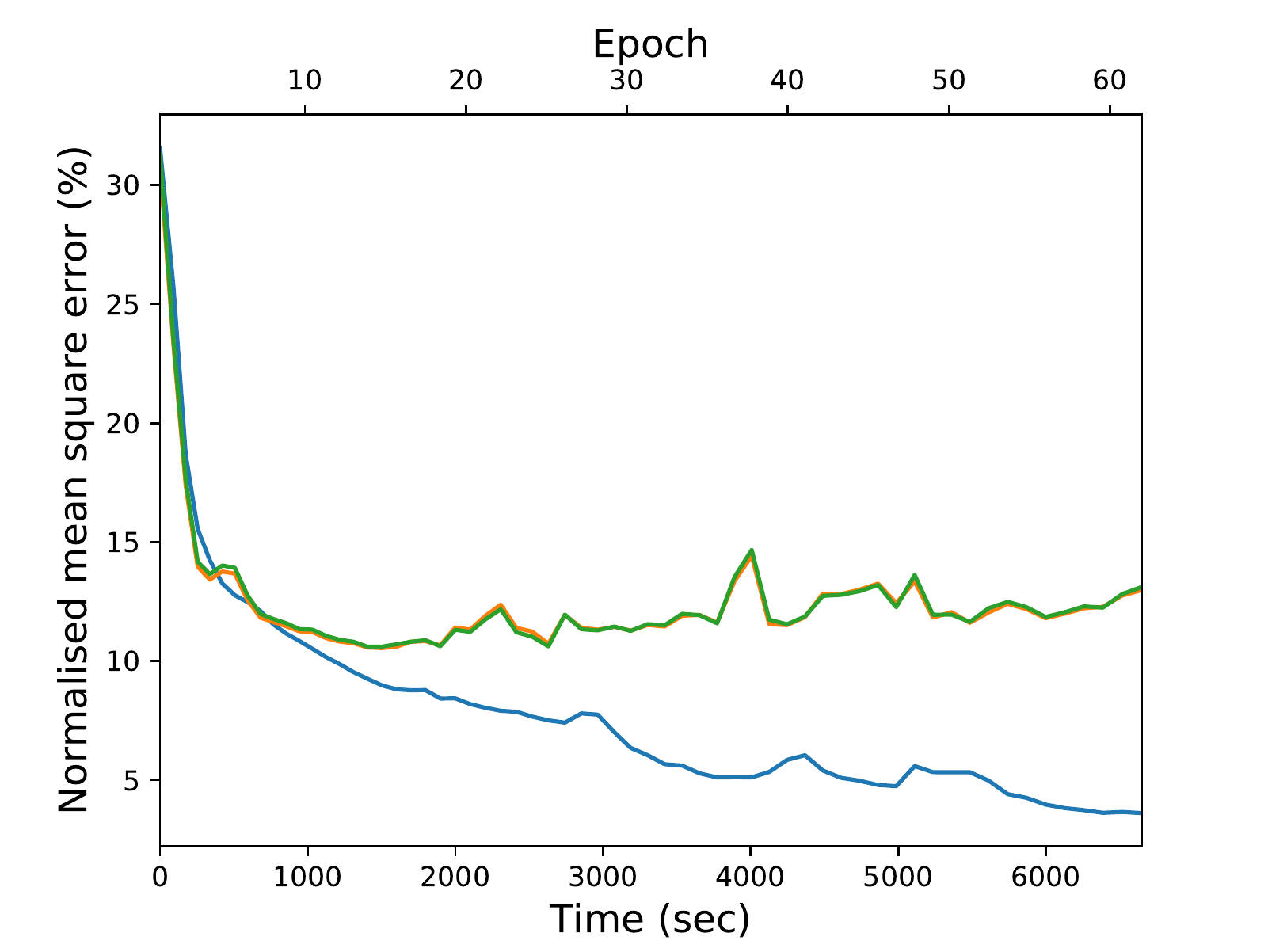}
    \caption{Training history of model using the augmented aggregative function, training (blue), validation (orange) and testing (green) datasets: first case study.}
    \label{fig:custom_aggr_training_history_no_temp_var}
\end{figure}

The first case study was repeated and the training history this time is shown in Figure \ref{fig:custom_aggr_training_history_no_temp_var}.  Because of the augmentation in
the aggregative function, an increase of inputs to the neural networks occurs and different sizes are selected by cross validation, as shown in Table\ref{Tab:GNN_arch_2}. An example of this increase in size in the input layer is clear in the global neural network of the second CB. The size following the previous approach of a simple mean
aggregative function was 120 units; this time it is 240 units, since all aggregations of features result in twice the number of attributes being fed into the later neural networks. For this run, the minimum was achieved much faster than before at around 18 epoches (1500 seconds = 25 minutes). The minimum errors were $10.54\%$ and $10.61\%$ on
the validation and testing sets respectively -- a considerable improvement. Again, the error is similar for both validation and testing datasets; after
extensive training (of the order of hours), the training data actually approached zero.

\subsection{Results: Case Study Two}
\label{subsec:res2}

\begin{figure}[htbp!]
    \centering
    \includegraphics[width=0.85\textwidth]{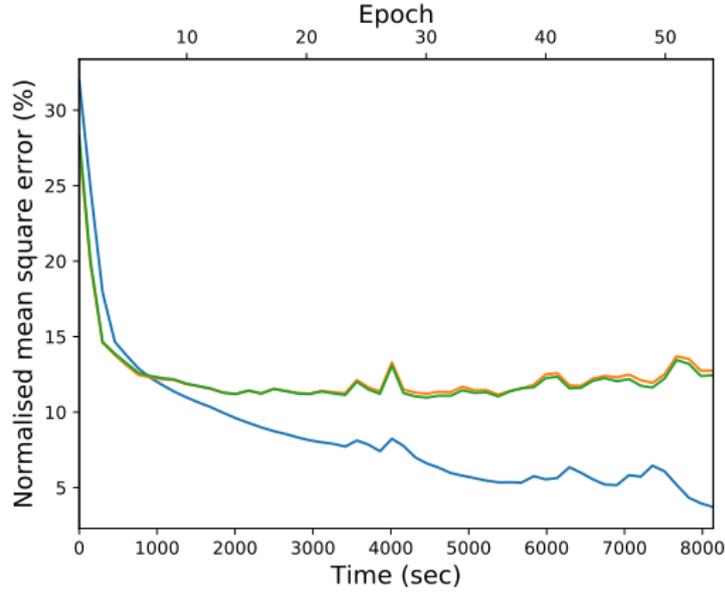}
    \caption{Training history of model using the augmented aggregative function, training (blue), validation (orange) and testing (green) datasets: second case study.}
    \label{fig:custom_aggr_training_history_one_member}
\end{figure}

For the second case study, the same approach was applied as before, but only with the augmented aggregative function. The training histories are shown in Figure \ref{fig:custom_aggr_training_history_one_member} and the network architectures in Table \ref{Tab:GNN_arch_3}. The convergence speed is of a similar order to that for
Case Study One. The minimum errors for validation and testing were 10.96\% and 11.21\% respectively. This is quite an impressive result, as the data now encompass
different temperatures for the trusses.

\begin{table}[htbp!]
\centering
\begin{tabular}{|l||l|l|l|}
\hline
		& First CB & Second CB & Third CB\\
\hline
	Edge NN     & 3, 64, 32 & 167, 80, 50   & 220, 180, 80 \\
\hline
	Node NN & 4, 72, 50  & 235, 100, 70  & 330, 300, 72 \\
\hline
	Global NN   & 1, 120, 85 & 325, 200, 100  & 404, 450, 150, 1 \\
\hline
\end{tabular}
\caption{Sizes of neural networks used in the GNN model with the augmented aggregative function: second case study.}
\label{Tab:GNN_arch_3}
\end{table}

\subsection{Results: Case Study Three}
\label{subsec:res3}

\begin{figure}[htbp!]
    \centering
    \includegraphics[width=0.85\textwidth]{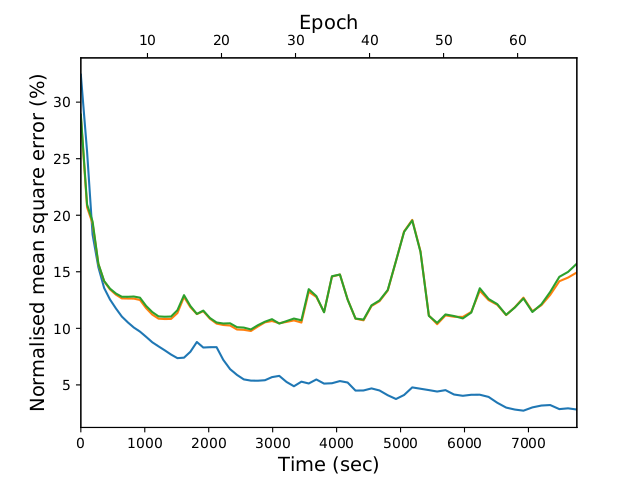}
    \caption{Training history of model using the augmented aggregative function, training (blue), validation (orange) and testing (green) datasets: third case study.}
    \label{fig:custom_aggr_training_history_two_members}
\end{figure}

Moving to the third case study and following the same approach, the training histories shown in Figure \ref{fig:custom_aggr_training_history_two_members} were obtained,
with network architectures as given in Table \ref{Tab:GNN_arch_4}. Edge input attributes this time also include the binary encoding of the member types, so there were five
instead of three. The validation and testing errors were 9.76\% and 9.90\% respectively. Again, this is good result;
in this case nonlinear temperature variation was added for a class of the members. So far, the algorithm appears to be actually improving with the complexity of the problem.

\begin{table}[htbp!]
\centering
\begin{tabular}{|l||l|l|l|}
\hline
		& First CB & Second CB & Third CB\\
\hline
	Edge NN     & 5, 100, 64 & 234, 100, 64   & 264, 250, 120 \\
\hline
	Node NN & 4, 120, 85  & 298, 200, 100  & 440, 450, 150 \\
\hline
	Global NN   & 1, 120, 85 & 413, 200, 100  & 640, 450, 150, 1 \\
\hline
\end{tabular}
\caption{Sizes of neural networks used in the GNN model with the augmented aggregative function: third case study.}
\label{Tab:GNN_arch_4}
\end{table}

In order to further test the extent to which the model learnt the underlying physics of estimating the natural frequency, it was also tested on larger trusses. The latter
dataset was created in the same way as in Case Study Three, but the number of nodes was randomly sampled in the interval $[41, 60]$. Interestingly enough and proving that
the algorithm has encoded the underlying physics satisfactorily, the test error on the larger truss dataset was 7.06\%.

Both of the later two case studies included temperature variations in the training data. This point is important, because it shows that the GNN approach also offers a
solution to the data normalisation problem discussed in Section \ref{sec:confound}. In order to do novelty detection across a population, one would need to have the
feature data for each structure at some reference temperature, otherwise the algorithm might signal novelty just because two undamaged structures produced features from
two different temperatures. The GNN algorithm solves this problem, because it can be probed to give the natural frequency corresponding to some reference temperature;
this is the normalisation step, also referred to as gauge fixing earlier in the paper.

\section{Discussion}
\label{sec:discussion}

In the first part of the paper, the objective was simply to set out some interesting geometrical structures -- differentiable manifolds and fibre bundles -- and frame some problems in population-based SHM in geometrical terms which may give insight, and thus point towards solutions. This part of the paper is quite speculative in its nature; it
may turn out to be the case that only the simplest SHM problems can be formulated in such abstract terms. However, it is certainly true to say that, in many cases where
problems in engineering and physics studied previously have been suited to a geometrical formulation, it has led to considerable insights into those problems.

The second part of the paper uses a machine learning algorithm to solve the normal section problem motivated by the geometrical approach. The GNN algorithm chosen, operates
directly on graph objects, and thus circumnavigates the graph embedding problem encountered where one attempts to map graph objects into vector spaces where classical machine
learning algorithms are most easily applied. The GNN algorithm is used to solve the normal section problem over a population of truss structures, allowing for temperature
variations across the training data for the algorithm. This is a challenging problem in PBSHM for at least two reasons: the first is that the population of structures is
heterogeneous, trusses can have widely-varying numbers of elements and disparate geometries. The second challenge comes from temperature variation, which can create the
problem of confounding influences even when one is conducting SHM for an individual structure.

The results of the exercise are very good and prove the potential of the GNN algorithm as a means of learning directly with attributed graphs, and in this case solve the
normal section problem for a natural frequency feature across a population of truss structures. An improved version of the GNN algorithm is demonstrated, based on an
augmentation of the aggregative functions used in the initial algorithm. The prediction error achieved in all three case studies shown here, is quite low,
especially considering how heterogeneous the structure population is. The algorithm seems to be learning the underlying physics of the problem; this belief is reinforced by the results here of applying the GNN model on larger trusses than the ones on which it was trained, where even lower errors were obtained on the prediction task.

\begin{figure}
\centering
\input{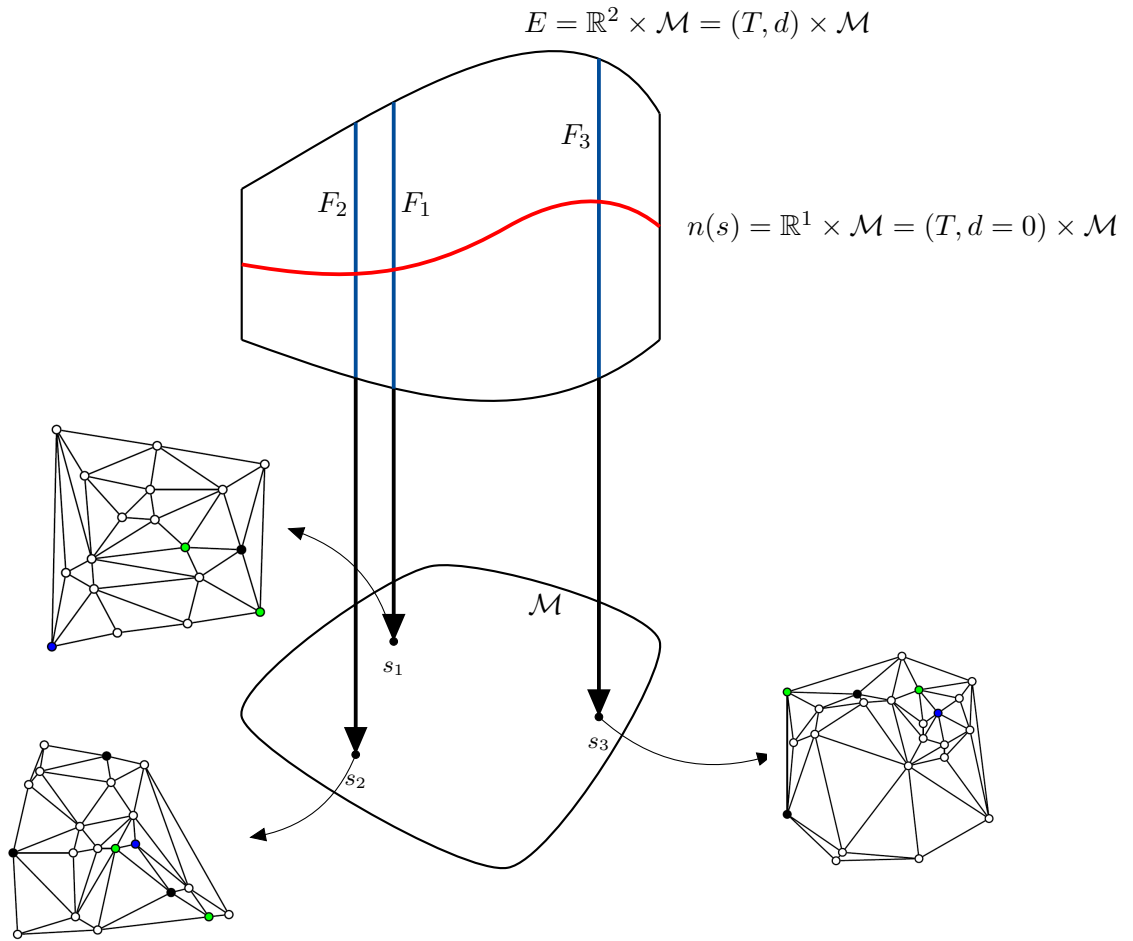}
\caption{Abstract scheme of the application. Nodes within $\mathcal{M}$ represent random trusses from within the population (3 examples shown here, black nodes represent
         fully fixed nodes, green nodes are fixed in the $x$ direction, blue are fixed in the $y$ direction and in white are free (pinned) nodes). A fibre $F_i$ for each
         structure $s_i$ is schematically shown as a blue line. It is parametrised by all potential first natural frequencies of the structure for varying temperature $T$
         and damage coefficient $d$. The algorithm here has approximated the normal condition cross section $n(s)$ where damage $d$ is equal to $0$ and for any potential temperature $T$}
    \label{fig:summary_geom}
\end{figure}

It is worth updating the general schematic in Figure \ref{fig:l9_AG_bundle} in terms of what was specifically achieved; the result is given in Figure \ref{fig:summary_geom}.
Although the base space is denoted $\mathcal{M}$, it is not a smooth manifold, it is a complex network; however, it is impossible here to show the detailed structure as
the training, validation and testing sets for the algorithm contained 16000 structures each. The normal section across the population has been estimated as desired; each
point on the section represents an undamaged structure at the reference temperature $T$. Another point is worth noting, which is that the GNN algorithm did not make
explicit use of the metric distance in the space of structures, it has interpolated within its own learnt embedding of the graphs, as encoded in its neural network
weights. When transfer learning is attempted, that metric distance will be important. As mentioned earlier, the assignment of a graph metric can be accomplished by a
graph matching network \cite{Li}.

It is important to emphasise why these results are interesting. They are {\em not} intended as a demonstration of how to estimate the natural frequency of trusses; this
problem is quite easily and accurately solved using Finite Element analysis. In practice, the PBSHM population of structures will be drawn from real experience, and it may span
a wide range of structural types from aeroplanes to wind turbines to bridges. In many cases, it would be expensive or otherwise infeasible to obtain FE models validated to
some appropriate level of confidence and accuracy. However, the PBSHM formulation proposes that SHM is carried out in a data-based framework with the structures represented
by Irreducible Element (IE) models and by Attributed Graphs (AGs). The exercise conducted here is intended to show that machine learning can be used directly on AGs in order
to solve a nontrivial PBSHM problem. While the truss structures used for illustration here are structurally quite simple, they are quite complicated in terms of their AG
representations, with some of the graphs considered here containing up to 60 nodes.

The potential use of categorical encoding has been applied here, and appears to be quite powerful, especially when the exact parameters required for an FE or other physics-based simulation, cannot be measured. In the truss example, characteristics such as the Young's modulus may have been measured in a laboratory by tension tests. These tests will have been performed within the highly-controlled environmental conditions of the laboratory, and not in the ones in which the part will be placed and operated. Another important
aspect is that a linear behaviour is usually assumed for these parameters, but may not apply in actual operation. For these reasons, the data-driven approaches can outmatch
physics-based approaches.

The GNNs are not totally `black-box models'. Inductive biases are an important novelty of the approach, whereby the knowledge of the user can be passed in the algorithm via
the structure of the data. The algorithm is much more flexible than classical machine learners, in that it can operate on general objects and encode their relations, it does
not require that input and output data occupy real vector spaces. It offers greater possibilities that the user can embed their prior knowledge of the objects of interest
and their relations. This flexibility makes the approach particularly well suited to geometrical and topological problems in structural dynamics, like the sort discussed
in the earlier part of the paper. It also means that the models have greater potential as `grey-boxes', augmenting the power of nonparametric learners with physical domain knowledge. In the case of PBSHM, a great deal of physical insight is used in forming the IE models; this is incorporated directly in the AG representations via their topology
and their attribute vectors. In the example presented here, it is clear that the GNN algorithm has captured the underlying physics, because the model has been successfully
applied to a set of larger trusses. This fact is evidence that the algorithm has {\em extrapolated} with some success, and this is a facility of physics-based models rather
than nonparametric black boxes. It is the subject of further research, but it is hoped that the geometrical framework discussed in the earlier part of the paper may lead to further inductive biases which could strengthen the GNN algorithm for the PBSHM problem.

\section{Conclusions}
\label{sec:conc}

The first part of this paper has outlined how some structures from differential geometry lend themselves to a theoretical basis for PBSHM. In particular, the formulation
of a collective of feature spaces as a bundle over a space of structures, is presented as a quite natural one; the problem of data normalisation then appears as a form of gauge-fixing. Transfer learning, which has been proposed separately as a basis for PBSHM \cite{PBSHMMSSP3} appears as a map between fibres in the bundle. While it is clear
that considerable research remains to be done in terms of a rigorous foundation, it is hoped that the geometrical approach will provide some insights along the way; not
least in terms of guiding inductive biases for machine learning, as discussed in the second part of the paper.

The second part of the paper looks at a concrete problem -- that of finding the normal cross section of the feature bundle over a heterogeneous population of structures.
This is a difficult and fundamental problem emerging from the geometrical viewpoint of PBSHM. Rather than a direct and pure geometrical attack on the problem, a machine
learning algorithm -- Graph Neural Networks (GNN) -- is motivated and applied. The potential power of this approach is demonstrated on a population of truss structures.

Finally, the GNN algorithm demonstrated here has not only solved the normal section problem. By learning the natural frequencies across the population, even in the presence
of the confounding influence of temperature, it has also solved the data normalisation problem for the context of interest.

\section*{Acknowledgements}

This project has received funding from the European Union’s Horizon 2020 research and innovation programme under the Marie Skłodowska-Curie grant agreement No 764547. KW would like to thank the UK Engineering and Physical Sciences Research Council (EPSRC) for an Established Career Fellowship (EP/R003645/1). C. Mylonas and E. Chatzi would further like to gratefully acknowledge the support of the European Research Council via the ERC Starting Grant WINDMIL (ERC-2015-StG \#679843).

\bibliographystyle{unsrt}
\bibliography{YMSSP_107692}

\begin{thebibliography}{10}

\bibitem{PBSHMMSSP1}
L.A.\ Bull, P.A.\ Gardner, J.\ Gosliga, T.J.\ Rogers, N.\ Dervilis, E.J.\
  Cross, E.\ Papatheou, A.E.\ Maguire, C.\ Campos, and K.\ Worden.
\newblock Foundations of population-based {SHM}, {P}art {I}: homogeneous
  populations and forms.
\newblock {\em Submitted to Mechanical Systems and Signal Processing}, 2020.

\bibitem{PBSHMMSSP2}
J.\ Gosliga, P.A.\ Gardner, L.A.\ Bull, N.\ Dervilis, and K.\ Worden.
\newblock Foundations of population-based structural health monitoring, part
  {II}: Heterogeneous populations -- graphs, networks and communities.
\newblock {\em Submitted to Mechanical Systems and Signal Processing}, 2020.

\bibitem{PBSHMMSSP3}
P.A.\ Gardner, L.A.\ Bull, J.\ Gosliga, N.\ Dervilis, and K.\ Worden.
\newblock Foundations of population-based {SHM}, {P}art {III}: heterogeneous
  populations -- transfer and mapping.
\newblock {\em Submitted to Mechanical Systems and Signal Processing}, 2020.

\bibitem{Schutz}
B.F.\ Schutz.
\newblock {\em Geometrical Methods of Mathematical Physics}.
\newblock Cambridge University Press, 1980.

\bibitem{Hamilton}
M.J.D.\ Hamilton.
\newblock {\em Mathematical Gauge Theory}.
\newblock Springer, 2017.

\bibitem{Alampalli}
S.\ Alampalli.
\newblock Effects of testing, analysis, damage, and environment on modal
  parameters.
\newblock {\em Mechanical Systems and Signal Processing}, 14:63--74, 2000.

\bibitem{FarrarI40}
C.R.\ Farrar, P.J.\ Cornwell, S.W.\ Doebling, and M.B.\ Prime.
\newblock Structural health monitoring studies of the {A}lamosa {C}anyon and
  {I}--40 {B}ridges.
\newblock Technical Report Los Alamos National Laboratory Report LA-13635-MS,
  2019.

\bibitem{Peeters}
B.\ Peeters and G.\~De Roeck.
\newblock One-year monitoring of the {Z24}-bridge: environmental effects versus
  damage events.
\newblock {\em Earthquake Engineering and Structural Dynamics}, 30:149--171,
  2001.

\bibitem{Sohn}
H.\ Sohn.
\newblock Effects of environmental and operational variability on structural
  health monitoring.
\newblock {\em Philosophical Transactions of the Royal Society A},
  365:539--560, 2007.

\bibitem{Steenrod}
N.\ Steenrod.
\newblock {\em The Topology of Fibre Bundles}.
\newblock Princeton University Press, 1951.

\bibitem{Husemoller}
D.\ Husemoller.
\newblock {\em Fibre Bundles}.
\newblock Springer, 1994.

\bibitem{Kobayashi}
S.\ Kobayashi and K.\ Nomizu.
\newblock {\em Foundations of Differential Geometry}, volume I, II.
\newblock Wiley Interscience, 1963.

\bibitem{Eguchi}
T.\ Eguchi, P.B.\ Gilkey, and A.J.\ Hanson.
\newblock Gravitation, gauge theories and differential geometry.
\newblock {\em Physics Reports}, 66:213--393, 1980.

\bibitem{Abraham}
R.\ Abraham and J.E.\ Marsden.
\newblock {\em Foundations of Mechanics}.
\newblock American Mathematical Society, 2008.

\bibitem{Reeb}
G.\ Reeb.
\newblock Vari{\'e}t{\'e}s symplectiques, vari{\'e}t{\'e}s presque-complexes et
  syst{\`e}mes dynamiques.
\newblock {\em Comptes Rendus de l'Acad\'emie des Sciences de Paris},
  235:776–--778, 1952.

\bibitem{Synge}
J.L.\ Synge.
\newblock On the geometry of dynamics.
\newblock {\em Philosophical Transactions of the Royal Society, Series A},
  220:31--106, 1926.

\bibitem{Mackey}
G.W.\ Mackey.
\newblock {\em Mathematical Foundations of Quantum Mechanics}.
\newblock Benjamin-Cummings, 1963.

\bibitem{Izenman}
A.J.\ Izenman.
\newblock {\em Modern Multivariate Statistical Techniques: Regression,
  Classification, and Manifold Learning}.
\newblock Springer, 2008.

\bibitem{McInnes}
L.\ McInnes, J.\ Healy, and J.\ Melville.
\newblock Umap: uniform manifold approximation and projection for dimension
  reduction.
\newblock {\em arXiv preprint:1802.03426v2 [stat.ML]}, 2018.

\bibitem{WordenNL}
K.\ Worden and G.R.\ Tomlinson.
\newblock {\em Nonlinearity in Structural Dynamics}.
\newblock Institute of Physics Press, 2001.

\bibitem{Birkhoff}
G.\ Birkhoff and S.\~Mac Lane.
\newblock {\em A Survey of Modern Algebra}.
\newblock Routledge, 2008.

\bibitem{Farrar}
C.R.\ Farrar and K.\ Worden.
\newblock {\em Structural Health Monitoring: A Machine Learning Perspective}.
\newblock John Wiley and Sons, 2011.

\bibitem{Mihaylov}
G.\ Mihaylov and M.\ Spallanzani.
\newblock Emergent behaviour in a system of industrial plants detected via
  manifold learning.
\newblock {\em Journal of Prognostics and Health Management}, 7, 2016.

\bibitem{WordenIEEE}
K.\ Worden, T.\ Baldacchino, J.\ Rowson, and E.J.\ Cross.
\newblock Some recent developments in {SHM} based on nonstationary time series
  analysis.
\newblock {\em Proceedings of the IEEE}, 106:1589--1603, 2016.

\bibitem{Cross1}
E.J.\ Cross, K.\ Worden, and Q.\ Chen.
\newblock Cointegration; a novel approach for the removal of environmental
  trends in structural health monitoring data.
\newblock {\em Proceedings of the Royal Society, Series A}, 467:2712--2732,
  2011.

\bibitem{Cross2}
E.J.\ Cross, K.\ Worden, G.\ Manson, and S.G.\ Pierce.
\newblock Features for damage detection with insensitivity to environmental and
  operational variations.
\newblock {\em Proceedings of the Royal Society, Series A}, 468:4098--4122,
  2012.

\bibitem{Manson}
G.\ Manson.
\newblock Identifying damage sensitive, environmental insensitive features for
  damage detection.
\newblock In {\em Proceedings of the 3$^{rd}$ International Conference on
  Identification in Engineering Systems, Swansea, UK}, 2002.

\bibitem{Kullaa}
J.\ Kullaa.
\newblock Is temperature measurement essential in {SHM}?
\newblock In {\em Proceedings of the 4$^{th}$ International Workshop on SHM,
  Palo Alto, CA.}, 2003.

\bibitem{Bunke}
H.\ Bunke and K.\ Shearer.
\newblock A graph distance metric based on the maximum common subgraph.
\newblock {\em Pattern Recognition Letters}, 19:255--259, 1998.

\bibitem{Fernandez}
M.-A.\ Fern\'{a}ndez and G.\ Valiente.
\newblock A graph distance metric combining maximum common subgraph and minimum
  common supergraph.
\newblock {\em Pattern Recognition Letters}, 22:753--758, 2001.

\bibitem{Li}
Y.\ Li, C.\ Gu, T.\ Dullion, O.\ Vinyals, and P.\ Kohli.
\newblock Graph matching networks for learning the similarity of graph
  structured objects.
\newblock Technical Report arXiv:1904.1278v2 [cs.LG], 2019.

\bibitem{Sun}
S.\ Sun, H.\ Shi, and Y.\ Wu.
\newblock A survey of multi-source domain adaptation.
\newblock {\em Information Fusion}, 24:84--92, 2015.

\bibitem{Bishop}
C.M.\ Bishop.
\newblock {\em Neural Networks for Pattern Recognition}.
\newblock Oxford University Press, 1995.

\bibitem{Bishop2}
C.M.\ Bishop.
\newblock {\em Pattern Recognition and Machine Learning}.
\newblock Springer-Verlag, 2006.

\bibitem{Bacciu}
D.\ Bacciu, F.\ Errica, A.\ Micheli, and M.\ Podda.
\newblock A gentle introduction to deep learning for graphs.
\newblock {\em Neural Networks}, 2020.

\bibitem{Cai}
H.\ Cai, V.W.\ Zheng, and K.C.\ Cheng.
\newblock A comprehensive survey of graph embedding: problems, techniques, and
  applications.
\newblock {\em IEEE Transactions on Knowledge and Data Engineering},
  30:1616--1637, 2018.

\bibitem{Battaglia}
P.W.\ Battaglia, J.B.\ Hamrick, V.\ Bapst, A.\ Sanchez{-}Gonzalez, V.\~Flores
  Zambaldi, M.\ Malinowski, A.\ Tacchetti, D.\ Raposo, A.\ Santoro, R.\
  Faulkner, C.\ G{\"{u}}l{\c{c}}ehre, H.F.\ Song, A.J.\ Ballard, J.\ Gilmer,
  G.E.\ Dahl, A.\ Vaswani, K.R.\ Allen, C.\ Nash, V.\ Langston, C.\ Dyer, N.\
  Heess, D.\ Wierstra, P.\ Kohli, M.\ Botvinick, O.\ Vinyals, Y.\ Li, and R.\
  Pascanu.
\newblock Relational inductive biases, deep learning, and graph networks.
\newblock (arXiv:1806.01261v3 [cs.LG]), 2018.

\bibitem{Cortes}
C.\ Cortes and V.\ Vapnik.
\newblock Support-vector networks.
\newblock {\em Machine Learning}, 20:273--297, 1995.

\bibitem{Perozzi}
B.\ Perozzi, R.\ Al-Rfou, and S.\ Skiena.
\newblock Deepwalk: {O}nline learning of social representations.
\newblock In {\em Proceedings of the 20$^{th}$ {AGM} {SIGKDD} International
  Conference on Knowledge Discovery and Data Mining}, pages 701--710, 2014.

\bibitem{Kipf}
T.N.\ Kipf and M.\ Welling.
\newblock Semi-supervised classification with graph convolutional networks.
\newblock {\em arXiv preprint:1609.02907}, 2016.

\bibitem{Zhang}
S.\ Zhang, H.\ Tong, J.\ Xu, and R.\ Maciejewski.
\newblock Graph convolutional networks: a comprehensive review.
\newblock {\em Computational Social Networks}, 6, 2019.

\bibitem{Dobson}
P.D.\ Dobson and A.J.\ Doig.
\newblock Distinguising enzyme structures from non-enzymes without alignments.
\newblock {\em Journal of Molecular Biology}, 330, 2003.

\bibitem{Ralaivola}
L.\ Ralaivola, S.J.\ Swadimas, H.\ Saigo, and P.\ Baldi.
\newblock Graph kernels for chemical informatics.
\newblock {\em Neural networks}, 18:1093--1110, 2005.

\bibitem{Zitnik}
M.\ Zitnik, M.\ Agrawal, and J.\ Leskovec.
\newblock Modeling polypharmacy side effects with graph convolutional neural
  network.
\newblock {\em Bioinformatics}, 34:i457--i466, 2018.

\bibitem{Wang}
X.\ Wang and A.\ Gupta.
\newblock Videos as space-time region graphs.
\newblock In {\em Computer Vision -- ECCV 2018}, pages 413--431, 2018.

\bibitem{Geron}
A.\ Ge{\'e}ron.
\newblock {\em Hands-on Machine Learning with Scikit-Learn, Keras, and
  TensorFlow: Concepts, Tools, and Techniques to Build Intelligent Systems}.
\newblock OReilly, second edition, 2019.

\bibitem{Delaunay}
B.\ Delaunay.
\newblock Sur la sph\`ere vide. a la m\'emoire de {G}eorges {V}oronoi.
\newblock {\em Bulletin de l'Acad\'emie des Sciences de l'URSS. Classe des
  Sciences Math\'ematiques et na}, 6:793–800, 1934.

\bibitem{Tarassenko}
L.\ Tarassenko.
\newblock {\em A Guide to Neural Computing Applications}.
\newblock Arnold, 1998.

\end{thebibliography}

\end{document}